\pdfoutput=1
\documentclass[11pt]{article}

\usepackage[final]{styles/acl}

\usepackage{times}
\usepackage{latexsym}

\usepackage[T1]{fontenc}
\usepackage[utf8]{inputenc}

\usepackage{microtype}

\usepackage{inconsolata}

\usepackage{graphicx}

\usepackage{amsmath}
\usepackage{float}
\usepackage{subcaption}

\usepackage{orcidlink}
\usepackage{booktabs}

\usepackage{algorithm}
\usepackage{algpseudocode}
\usepackage{tabularx}

\title{Complexity-aware fine-tuning}

\author{
 \textbf{Andrey Goncharov\textsuperscript{1}\orcidlink{0009-0006-9471-2181}},
 \textbf{Daniil Vyazhev\textsuperscript{1}\orcidlink{0009-0004-4521-9062}},
 \textbf{Petr Sychev\textsuperscript{1}},\\
 \textbf{Edvard Khalafyan},
 \textbf{Alexey Zaytsev\textsuperscript{1}\orcidlink{0000-0002-1653-0204}},
\\
 \textsuperscript{1}Applied AI Institute
\\
}

\begin{document}
\maketitle

\begin{abstract}
% (0.5/0.6 vs. 0.4/0.46 accuracies across 2 models)
% We also provide an in-depth analysis of alternative complexity estimation techniques based on expert assessment via model-as-judge (MASJ) and chain-of-thought entropy aggregation with ROC AUC scores of 0.57 and 0.7 accordingly. 

General-purpose Large Language Models (LLMs) are frequently fine-tuned through supervised fine-tuning (SFT) to enhance performance in specific domains. Better results can be achieved by distilling the chain-of-thought of a larger model at the cost of numerous expensive calls and a much greater amount of data.
We propose a novel blueprint for efficient fine-tuning that uses reasoning only for complex data identified by entropy. Specifically, across three small open models ($\approx 3B$) we split the training data into complexity categories by a single token answer entropy (ROC AUC $0.73$), fine-tune large language models (LLMs) via SFT and distillation, and show that our pipeline significantly outperforms the standard SFT approach ($0.58$ vs $0.45$ average accuracy) and outperforms the distillation approach ($0.58$ vs $0.56$ average accuracy) while using $81\%$ less data.
We publish our code\footnote{\url{https://github.com/LabARSS/complexity-aware-fine-tuning}} and data\footnote{\url{https://github.com/LabARSS/complexity-aware-fine-tuning?tab=readme-ov-file\#data}} to facilitate further research in this direction.
\end{abstract}

% General purpose Large Language Models (LLMs) are frequently fine-tuned to improve performance in niche domains. Although fine-tuning is a standard practice, we still lack a deep understanding of how to aggregate data for better results. In this work, we show that the entropy-based output estimation provides a meaningful guideline for fine-tuning data preparation. Specifically, across two small open models ($\approx 3B$) we find that a single token answer entropy shows ROC AUC score of $\approx 0.73$ and allows us to split the training data into three complexity categories to apply different tuning mechanisms. As result, we propose a novel blueprint for efficient fine-tuning that outperforms the standard approach $0.55$ vs $0.43$ average accuracy. Our findings show immediate enhancements in fine-tuning performance. We publish our code\footnote{\url{https://github.com/LabARSS/complexity-aware-fine-tuning}} and data\footnote{\url{https://huggingface.co/datasets/LabARSS}} to facilitate further investigation and immersion of the numerical complexity analysis.

\section{Introduction}

General-purpose LLMs excel across diverse tasks, but deploying them is often impractical under constraints on compute, latency or cost. These realities motivate compact, domain-adapted models that can deliver competitive or superior performance. 
Growing evidence shows that carefully tuned smaller models can match or outperform larger open models in mathematics \cite{yang2024qwen25mathtechnicalreportmathematical}, medicine \cite{Wu2025}, chemistry \cite{yu2024llasmol}.
Thus, smaller domain-specific LLMs appear a compelling choice, especially under resource constraints.

A standard way for training is domain-adaptive fine-tuning: starting from a general-purpose LLM and training it on in-domain data \cite{parthasarathy2024ultimate} with optional distillation from a larger model. Recent work has mainly concentrated on optimizing training mechanics (objectives, schedulers, and regularizers), whereas comparatively little attention has been paid to data curation. Curriculum-style approaches have also been explored \cite{kim2024strategic, shi2025efficient} with increasing complexity of data during training, but with limited gains in final accuracy and efficiency, suggesting that more principled strategies are needed.

\begin{figure}
    \centering
    \includegraphics[width=1\linewidth]{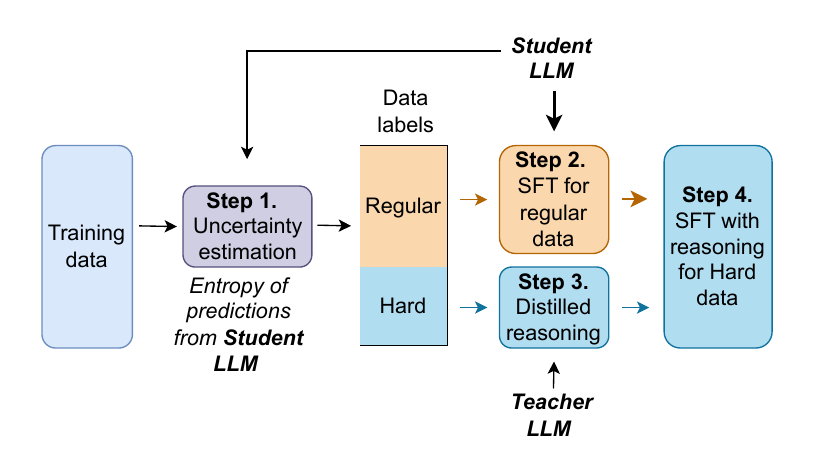}
    \caption{Complexity-aware fine-tuning scheme for a student LLM: we identify complexity of questions via uncertainty estimation of a model (Step 1), then for questions of regular complexity we apply direct SFT (Step 2), while for hard questions we include reasoning from a teacher LLM (Step 3) and complete SFT using reasoning-enriched hard data (Step 4).}
    \label{fig:reasoning_fine_tune_scheme}
\end{figure}

To address this gap, we propose a fully automated pipeline, presented in  Figure~\ref{fig:reasoning_fine_tune_scheme}, that focus on both accuracy and efficiency of the fine-tuning procedure. 
It consists of two steps: (1) a split of the data for fine-tuning by the complexity of the task to regular and hard questions and (2) training a model via supervised fine-tuning (SFT) followed by another round of SFT that utilizes distillation on a chain-of-thought from a larger model \cite{li-etal-2023-symbolic, hsieh2023distillingstepbystepoutperforminglarger}. 
We confirm the effectiveness of the framework by fine-tuning three open models: Qwen2.5-3B \cite{qwen2}, Phi-4-Mini \cite{microsoft2025phi4mini}, LLama 3.2 3B \cite{grattafiori2024llama3herdmodels}, on three datasets: multiple choice question answering dataset MMLU-Pro, multiple choice question answering dataset MedMCQA, mathematical dataset GSM8K.

For both steps of the pipeline, we provide sensitivity studies to the design choices made.
In step one, we consider different options for complexity estimation based on a model's confidence in its answer. 
Our detailed study considers a wide range of methods, including model-as-judge (MASJ), entropy-based aggregation, and calculation methods in line with findings from \citet{fadeeva2023lmpolygraphuncertaintyestimationlanguage} with a top performing one being an answer entropy.
In step two, we explore training on different complexity subsets of the distilled chain-of-thought --- concluding that for regular data standard SFT is enough, while for hard data we benefit reasoning-based distillation.
This observation constrains the reasoning to hard data only, making our approach token-efficient. 

The specific contributions are the following:
\label{sec:contributions}
\begin{itemize}
    \item A training pipeline to fine-tune LLMs via SFT and distillation for regular and hard data correspondingly. Our procedure provides accuracies: 
    \begin{itemize}
        \item (MMLU-Pro) $0.52$ and $0.64$ c.t. $0.39$ and $0.51$ for SFT and $0.50$ and $0.63$ for distillation baselines.
        \item (GSM8K) $0.82$ and $0.89$ c.t. $0.13$ and $0.19$ for SFT and $0.79$ and $0.89$ for distillation baselines.
        \item (MedMCQA) $0.56$ and $0.58$ c.t. $0.51$ and $0.51$ for SFT and $0.56$ and $0.57$ for distillation baselines.
    \end{itemize}
    \item An unsupervised method to split a multiple-choice question answering dataset by complexity based on the token-wise entropy of the response with ROC AUC $0.73$. Alternative complexity estimators rooted in MASJ and aggregated entropy, augmented with reasoning results and various aggregation techniques yield inferior results.
    \item Open-source standardized datasets %\footnote{\url{https://github.com/LabARSS/complexity-aware-fine-tuning?tab=readme-ov-file\#data}} 
    to facilitate further development of uncertainty estimation and calibration methods: with and without chain-of-thought, with token probability distribution at each step provided, as well as additional~scores.
\end{itemize}

\section{Related works}

% distillation in a black box setting

% usage of complexity of questions or domains for more efficient fune-tuning or training

% compexity estimation?

\paragraph{Data complexity-aware learning.}
Curriculum learning has been explored to improve LLM fine-tuning by ordering training examples from easy to hard. \citet{kim2024strategic} propose sorting fine-tuning data by difficulty metrics (e.g. prompt length, model attention scores, and initial loss) so that the model learns on simpler prompts before complex ones. They found that this curriculum strategy yielded slightly higher accuracy than random shuffling, with ordering by an attention-based criterion performing best. This approach is attractive because it boosts performance without adding more data or parameters. However, the gains were modest, and defining difficulty automatically can be tricky - their method requires measuring things like loss or attention per example. 

Another strategy is filtering training data for quality. A notable example is LIMA \cite{zhou2023lima}, which shows that a large pre-trained model can be fine-tuned on just a small, high-quality subset of data. They fine-tuned a 65B Llama model on only 1000 carefully curated prompt-response pairs (chosen for diversity and clarity) without any reinforcement learning. Despite the tiny dataset, the resulting model performed remarkably well, learning to handle complex queries and even generalizing to tasks not seen in training. In a human evaluation, LIMA's answers were preferred over GPT-4's in 0.43 of cases. This "less is more" result suggests that much of an LLM's ability comes from pre-training, and fine-tuning needs only a small amount of exemplary data to unlock it. However, LIMA relied on a large base model and manual data curation. The approach may not scale down to smaller models and requires human intervention.

Another notable example of curated data selection is the SmallToLarge (S2L) method by \citet{yang2024smalltolarges2lscalabledata}, which leverages training trajectories from small models to guide the data selection for larger models. This way, the large LLM is trained on a diverse yet compact dataset covering different difficulty levels. S2L showed impressive results: for a math word problem dataset, they achieved the same accuracy using only 11\% of the data, and even outperformed other selection methods by ~4.7\% on average across several benchmarks. The strength of this approach is that it makes complexity-based data filtering automated and cheap. One caveat is the extra step of training a smaller model and clustering. The approach is mostly tested on specialized domains (math problems, clinical text summarization), so its generality to all types of tasks needs further validation. Additionally, it requires a large amount of data to make a filtered subset.

\paragraph{Complexity estimation.}
\citet{fadeeva2023lmpolygraphuncertaintyestimationlanguage} provides an in-depth comparison of multiple black-box and white-box methods of complexity estimation. They present promising results for white-box methods rooted in sequence probability and entropy aggregation. \citet{sychev2025llmapprehensiveanswers} focus specifically on entropy-based aggregations augmented with model-as-judge (MASJ) categorization by a reasoning score. They confirm that LLM's token-level entropy of the output is a good predictor of question difficulty, especially in knowledge-based domains. They also introduce MASJ reasoning score to estimate the question complexity. However, the authors use these metrics only to analyze model behavior. They do not integrate it into a practical data aggregation or fine-tuning workflow.

\paragraph{Research gap.} The question if we can combine novel complexity estimation techniques based on entropy aggregates with adaptive learning methods remains open. It is also unclear how we can use the insights for the reasoning estimates to dynamically change our learning approach. This paper aims to address these gaps.

\section{Methods}

\subsection{Training pipeline}
\label{sec:pipeline}

We propose the complexity-aware fine-tuning pipeline detailed in Figure \ref{fig:reasoning_fine_tune_scheme} with the following major stages: complexity estimation, data aggregation, fine-tuning.

\paragraph{1. Complexity estimation.}
We adopt the entropy of the answer token in the response as our primary complexity metric. We prompt the model to pick the correct option directly, without producing a chain-of-thought. See Section~\ref{sec:method_single_token_entropy} for details.

\paragraph{2. Data aggregation.}
To aggregate the data into groups by complexity (regular, hard), we evenly divide the dataset into three parts, ordering the entries by entropy of the response. A group with lower entropy values is categorized as easy and with the higher values as hard.
% , as Figure \ref{fig:reasoning_fine_tune_data_split} depicts.

\paragraph{3. Fine-tuning.}
We propose applying different fine-tuning strategies based on the complexity of the data.

For the regular groups, we use vanilla SFT \cite{howard2018universallanguagemodelfinetuning, raffel2023exploringlimitstransferlearning}, an established and robust practice. 
It involves fine-tuning a pretrained LM on labeled examples using a standard supervised objective, specifically, cross-entropy.
Used prompts are provided in Table~\ref{tab:prs_uqs_uq} in Appendix. 

As to the hard group, we hypothesize that hard questions require multiple logical steps and multiple attempts to learn core facts.
So, the standard SFT is suboptimal. 
We propose to elicit a chain-of-thought and allow the model to incrementally build the answer step-by-step as suggested by \citet{wei2023chainofthoughtpromptingelicitsreasoning}.

In this work, we apply the distillation technique, which involves training a smaller student model on the chain-of-thought of a larger LLM. It is a well-known practice supported by \cite{hsieh2023distillingstepbystepoutperforminglarger}. 
To create the distillation training samples, we prompt a large LLM to answer the multiple-choice question and produce a chain-of-thought in the process. Next, the whole response is attached to the dataset and used to train the smaller model. 

Pseudo-code for the algorithm is provided below in Algorithm~\ref{alg:complexity-aware-ft}.
% prompt table is missing: (\ref{tab:prompt_cot_response}) 

\begin{algorithm}
\caption{Complexity-Aware Fine-Tuning Pipeline}
\label{alg:complexity-aware-ft}
\begin{algorithmic}[1]
\Require Training dataset $\mathcal{D} = \{(q_i, a_i)\}_{i=1}^{N}$, Student LLM $\mathcal{M}_S$, Teacher LLM $\mathcal{M}_T$
\Ensure Fine-tuned Student LLM $\mathcal{M}_S^*$

\Statex \textbf{// Step 1: Uncertainty Estimation}
\For{each $(q_i, a_i) \in \mathcal{D}$}
    \State $\mathbf{p}_i \gets \mathcal{M}_S(q_i)$ \Comment{Get token probability distribution}
    \State $h_i \gets -\sum_{j=1}^{|\mathcal{V}|} p_{ij} \log p_{ij}$ \Comment{Compute answer token entropy}
\EndFor

\Statex
\Statex \textbf{// Step 2: Data Aggregation by Complexity}
\State Sort $\mathcal{D}$ by entropy values $\{h_i\}_{i=1}^{N}$ in ascending order
\State $\mathcal{D}_{\text{easy}} \gets \mathcal{D}[1 : \lfloor N/4 \rfloor]$ \Comment{Low entropy = easy}
\State $\mathcal{D}_{\text{medium}} \gets \mathcal{D}[\lfloor N/4 \rfloor + 1 : \lfloor 3N/4 \rfloor]$
\State $\mathcal{D}_{\text{hard}} \gets \mathcal{D}[\lfloor 3N/4 \rfloor + 1 : N]$ \Comment{High entropy = hard}
\State $\mathcal{D}_{\text{regular}} \gets \mathcal{D}_{\text{easy}} \cup \mathcal{D}_{\text{medium}}$

\Statex
\Statex \textbf{// Step 3: Standard SFT on Regular Data}
\For{epoch $= 1$ to $E_1$}
    \For{batch $\mathcal{B} \subset \mathcal{D}_{\text{regular}}$}
        \State $\mathcal{L}_{\text{SFT}} \gets -\sum_{(q,a) \in \mathcal{B}} \log P_{\mathcal{M}_S}(a \mid q)$
        \State $\mathcal{M}_S \gets \mathcal{M}_S - \eta \nabla \mathcal{L}_{\text{SFT}}$
    \EndFor
\EndFor

\Statex
\Statex \textbf{// Step 4: Distill Chain-of-Thought for Hard Data}
\For{each $(q_i, a_i) \in \mathcal{D}_{\text{hard}}$}
    \State $r_i \gets \mathcal{M}_T(q_i)$ \Comment{Generate CoT reasoning from teacher}
    \State $\mathcal{D}_{\text{hard}}^{\text{CoT}} \gets \mathcal{D}_{\text{hard}}^{\text{CoT}} \cup \{(q_i, r_i, a_i)\}$
\EndFor

\Statex
\Statex \textbf{// Step 5: SFT with Reasoning on Hard Data}
\For{epoch $= 1$ to $E_2$}
    \For{batch $\mathcal{B} \subset \mathcal{D}_{\text{hard}}^{\text{CoT}}$}
        \State $\mathcal{L}_{\text{CoT}} \gets -\sum_{(q,r,a) \in \mathcal{B}} \log P_{\mathcal{M}_S}(r \oplus a \mid q)$
        \State $\mathcal{M}_S \gets \mathcal{M}_S - \eta \nabla \mathcal{L}_{\text{CoT}}$
    \EndFor
\EndFor

\State \Return $\mathcal{M}_S^* \gets \mathcal{M}_S$
\end{algorithmic}
\end{algorithm}

\subsection{Complexity estimation approaches}

To find the most suitable metric for the training pipeline, we analyze the performance of the following techniques:
a single token answer entropy,
a chain-of-thought answer entropy,
a chain-of-thought aggregated response entropy, 
MASJ reasoning score, and
MASJ education level.
Used prompts are available in Appendix~\ref{sec:uq_prompts}.

\subsubsection{Answer entropy}
\label{sec:method_single_token_entropy}

\paragraph{Single token answer entropy.} In a similar fashion to that proposed by \cite{kadavath2022language}, we calculate the entropy of the answer token in the response.  The assumption is that the response uncertainty is a natural predictor of the question complexity. We prompt the model to answer the question directly (as a single token) and calculate token-wise entropy of the response as follows:
$$
h = -\sum_{i = 1}^{n} p_i \log{p_i},
$$
where $p_i$ is the probability of a single token and $n$ is the vocabulary size.
This is the main method used in our pipeline.

Additionally, similarly to \cite{zhou-etal-2023-context}, we examine the performance when we allow the model to explicitly say "I do not know" (IDK).

\paragraph{Chain-of-thought answer entropy.}
With the same assumption as for the single token entropy, we analyze the entropy of the answer token, but change the prompt to elicit a chain-of-thought type of response. The hypothesis is that, through the chain-of-thought process, the LLM can incrementally accumulate entropy, resulting in a better separation of certain and uncertain answers.

\subsubsection{Chain-of-thought aggregated response entropy.}
Building upon the single-token entropy approach, we investigate more sophisticated methods in line with \citet{fadeeva2023lmpolygraphuncertaintyestimationlanguage} for complexity estimation by analyzing the entire chain-of-thought (CoT) response. 
While the answer token entropy provides a localized measure of uncertainty, aggregating entropy across the complete reasoning process potentially offers a more comprehensive complexity assessment.

We evaluate ten distinct aggregation methods applied to CoT responses, comparing their effectiveness through ROC AUC metrics across multiple models, with and without "I don't know" option in the prompt. 
We consider word aggregation, sequence aggregation, and probability-based methods. See Appendix \ref{sec:aggr_methods} for details.

\subsubsection{MASJ reasoning score}

As one of the expert-like metrics, we ask a large LLM to estimate the number of logical steps required to answer the question. The hypothesis is that the questions that require more reasoning should be harder for the model to answer.

To collect the MASJ-based reasoning score, we go over the multiple-choice question answering dataset and query a large auxiliary LLM for the estimate. We prompt the model to provide the number of logical steps required to answer the question: low, medium, or high. Next, we query the large LLM again to estimate the quality of the previous assessment from 1 to 10, following the practice introduced in MT-Bench by \citet{zheng2023judging}. It allows us to filter out low quality scores by keeping only the ones with rates above or equal to 9.

\subsubsection{MASJ education level}

Like the other expert-like metrics, we ask a large LLM to estimate the required level of education to answer the question correctly. It is a natural human-like value used in other datasets \cite{rein2023gpqagraduatelevelgoogleproofqa, lu2022learn}.

We follow the same procedure as for the MASJ reasoning score, but use a different prompt.

% \subsubsection{Thinking and answer statistics of reasoning model}

% To further investigate how numerical complexity estimates can be applied for uncertainty quantification, we analyze the entropy and length of the reasoning chain for the current state-of-the-art (SOTA) reasoning model. 

% During inference, for each newly generated token, we store the probability distribution over the vocabulary of tokens with non-zero probabilities. To find the importance of the features, we train a logistic regression classifier using the scikit-learn \cite{scikit-learn} to predict the correctness of the model answer.

\section{Results}

\subsection{Experimental setup}

% likzet: good
\paragraph{Datasets}
We conduct experiments across multiple datasets: 

\begin{itemize}
    \item Multiple choice question answering dataset MMLU-Pro~\cite{wang2024mmluprorobustchallengingmultitask},  widely adopted by the community as one of the main performance benchmarks. 
    It spans 14 domains, offering a broad selection of questions of varying complexities. 
    Each question has approximately ten options, with a single correct one, which eliminates ambiguity in evaluation.
    \item GSM8K (Grade School Math 8K)~\cite{cobbe2021gsm8k} - dataset of 8.5K high quality linguistically diverse grade school math word problems.
    \item MedMCQA~\cite{pmlr-v174-pal22a} - large-scale, Multiple-Choice Question Answering (MCQA) dataset designed to address real-world medical entrance exam questions. It has more than 194k high-quality AIIMS and NEET PG entrance exam MCQs covering 2.4k healthcare topics and 21 medical subjects are collected with an average token length of 12.77 and high topical diversity
\end{itemize}

\paragraph{LLMs}
Regarding the models \footnote{All models and datasets are published under permissive licenses that allow them to be used for research purposes}, we utilize a variety of open models to analyze how the trend changes with model size. 
At the same time, our focus is on smaller models for fine-tuning to make our results reproducible and more relevant to the distillation scenario.

We apply our pipeline to three models: 
Qwen2.5-3B, Phi-4-Mini, LLama 3.2 3B.
For them, we measure single token and chain-of-thought entropy, collect other response metadata, fine-tune the models, and evaluate the overall pipeline.

Bigger models are used to further extend the entropy aggregate and MASJ analysis: Mistral 24B, Phi-4, Mistral 123B \cite{mistral123b}.
For the chain-of-thought distillation we utilize an ensemble of models: DeepSeek-V3-0324 \cite{deepseekai2024deepseekv3technicalreport}, Qwen 3 235B, Llama 4 Maverick \cite{llama4}.

\subsubsection{Our method and baselines}

In experiments, we performed training for 10 or 20 epochs; see more information below.
In the case of the reasoning used during training, we split the training into two equal halves for the first and the second round of fine-tuning.

\paragraph{Pipeline (ours)} For easy and medium groups, we perform SFT for five epochs. For the hard group, we apply learning from a distilled chain-of-thought of a larger model for another five epochs.

\paragraph{Alternative} As the alternative approach, we perform SFT for five epochs for the hard group. Next, for easy and medium groups, we apply learning from a distilled chain-of-thought of a larger model for another five epochs.

\paragraph{SFT} As our first baseline, we train the model via SFT without the data split for 10 epochs. 

\paragraph{Curriculum} For the second baseline, we train the model via curriculum-based SFT: 3 epochs on easy data, 3 epochs on medium data, and 4 epochs on hard data. 

\paragraph{Distillation} As an idealistic target, we tuned the LLM via distillation without the data split for 10 epochs.

\subsubsection{Technical details}

Based on ROC AUC results and SFT's positive performance on medium and easy questions (see \ref{sec:sensitivity_study}), we use single-token entropy as our complexity metric for the pipeline described in Section~\ref{sec:pipeline}. 

We randomly split the data into train, validation, and test with a ratio of 70:10:20. Next, in each chunk, we balance the data so that the number of entries in each complexity group is equal. 
Since there are fewer hard questions, we filter out medium and easy ones to match the size of the hard group.

\subsection{Main results}
\label{sec:pipeline_results}

\begin{figure*}[ht]
    \centering
    \begin{subfigure}[b]{0.48\linewidth}
        \centering
        \includegraphics[width=1\linewidth]{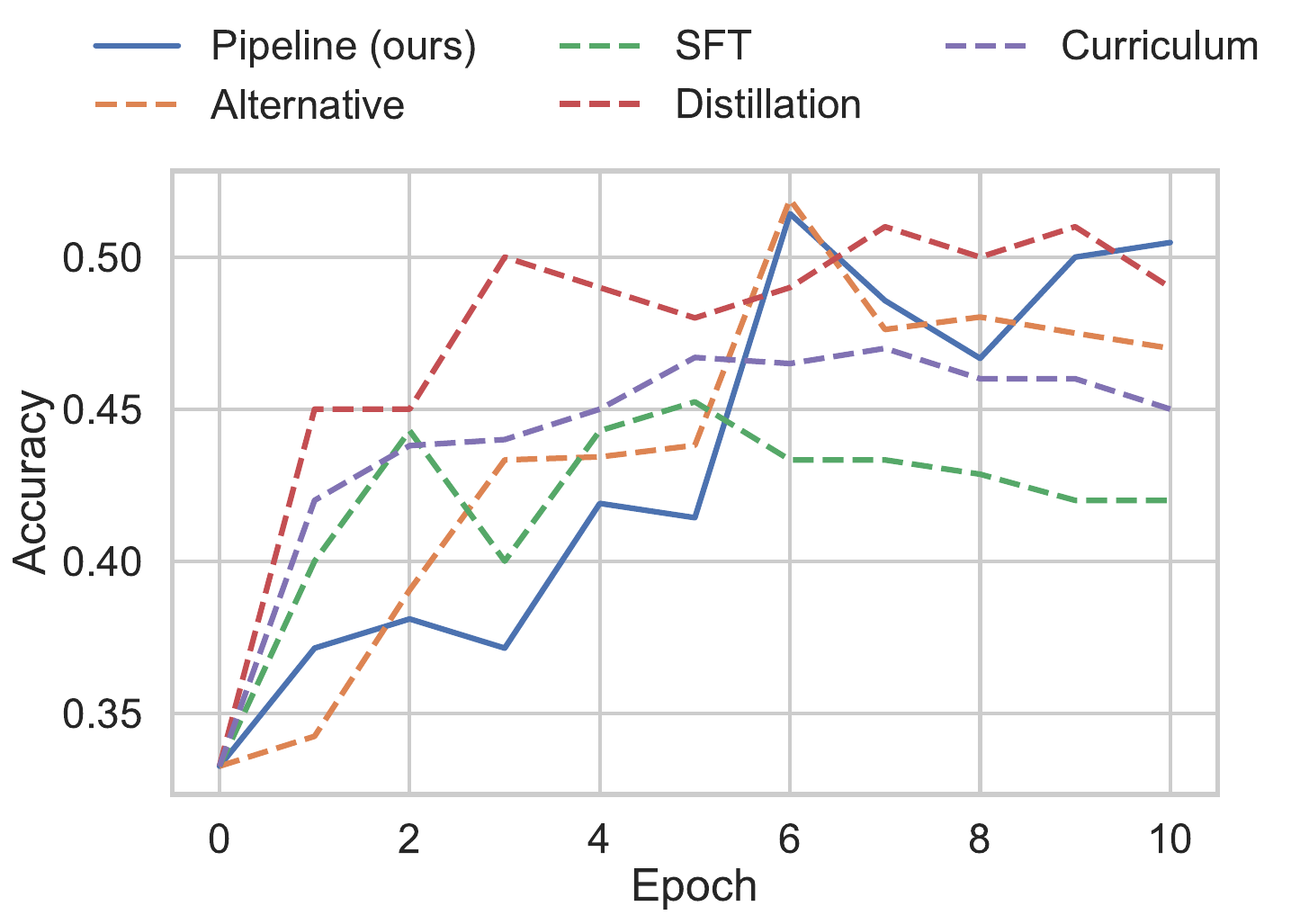}
        \caption{Accuracy for fine-tuning pipelines after 10 epochs for Qwen 3B (MMLU-Pro)}
        \label{fig:qwen_pipeline}
    \end{subfigure}
    \hfill
    \begin{subfigure}[b]{0.48\linewidth}
        \centering
        \includegraphics[width=1\linewidth]{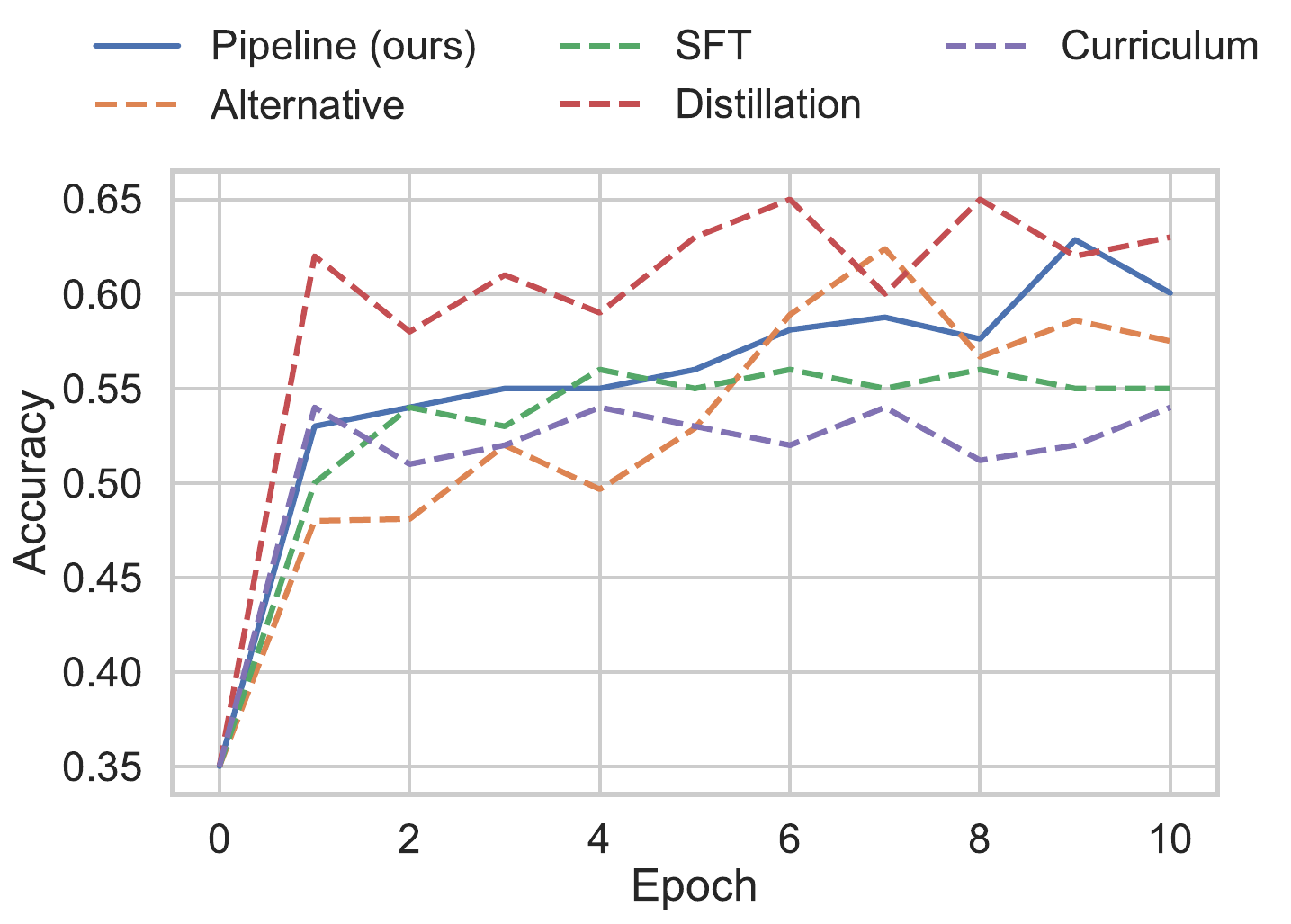}
        \caption{Accuracy for fine-tuning pipelines after 10 epochs for Phi-4-mini (MMLU-Pro)}
        \label{fig:phi_pipeline}
    \end{subfigure}
\end{figure*}
\begin{figure*}[h]
    \centering
    \begin{subfigure}[b]{0.48\linewidth}
        \centering
        \includegraphics[width=1\linewidth]{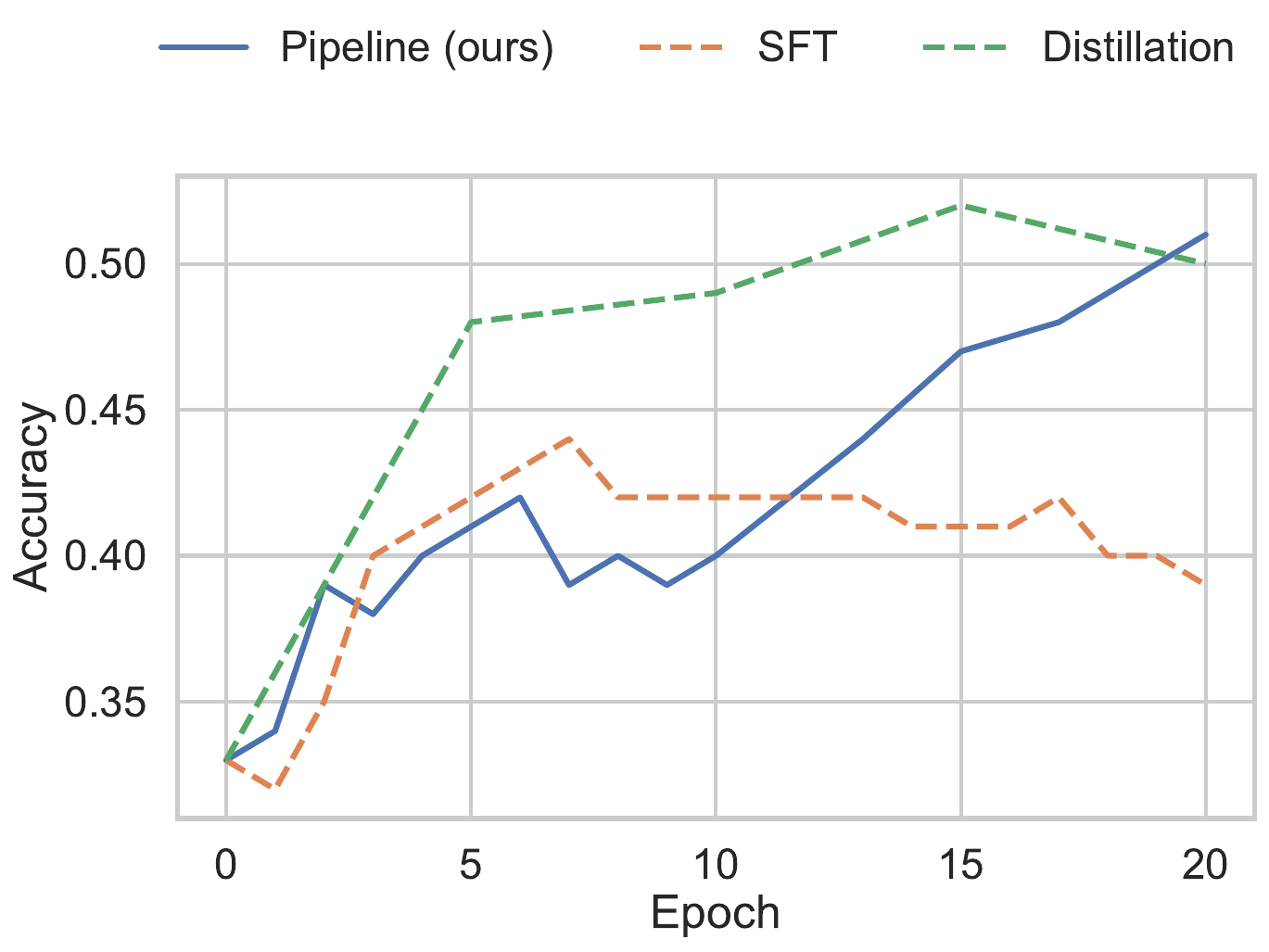}
        \caption{Accuracy for fine-tuning pipelines after 20 epochs for Qwen 3B (MMLU-Pro)}
        \label{fig:qwen_pipeline_20epochs}
    \end{subfigure}
    \hfill
    \begin{subfigure}[b]{0.48\linewidth}
        \centering
        \includegraphics[width=1\linewidth]{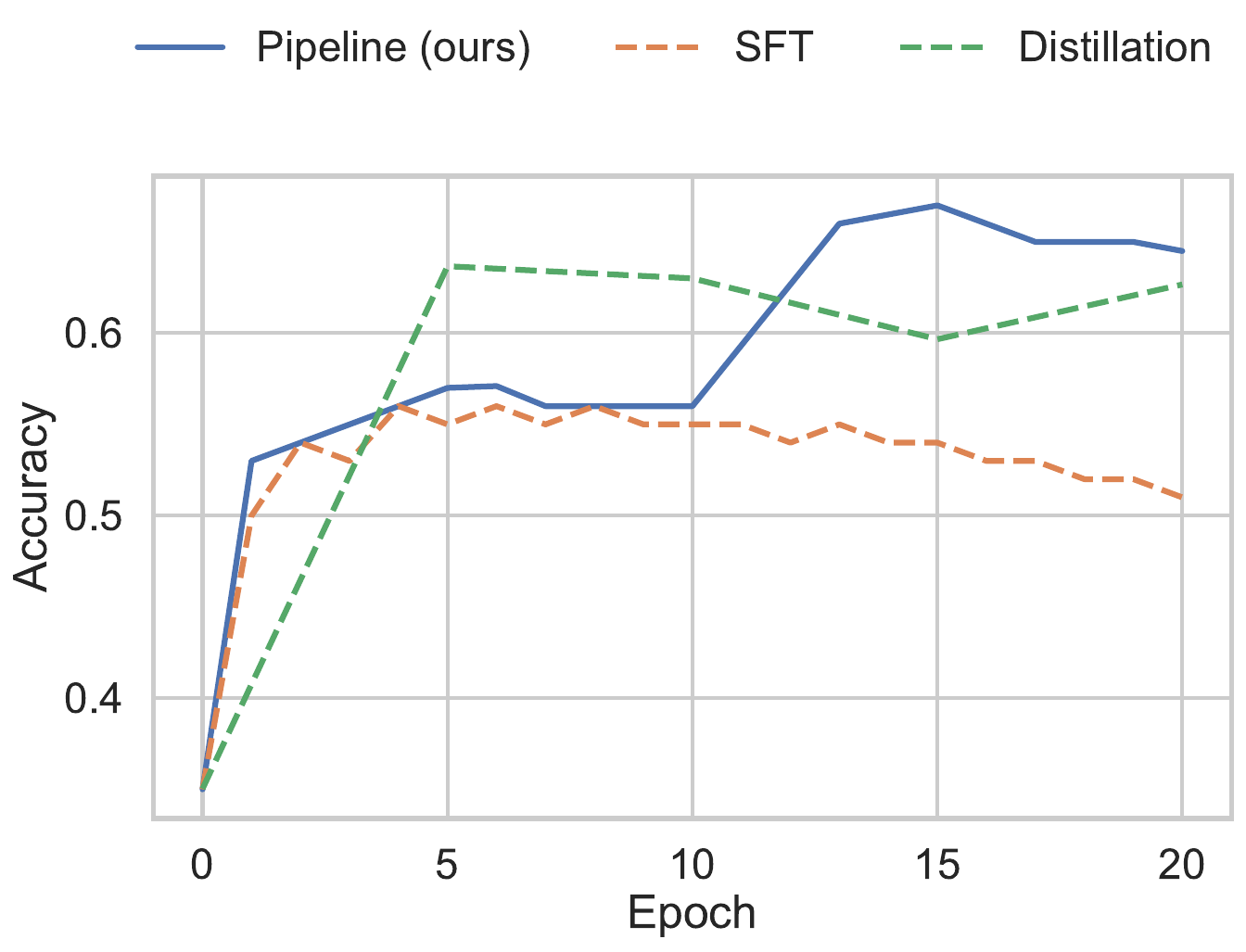}
        \caption{Accuracy for fine-tuning pipelines after 20 epochs for Phi-4-mini (MMLU-Pro)}
        \label{fig:phi_pipeline_20epochs}
    \end{subfigure}
\end{figure*}

Table ~\ref{tab:results_10epochs} and Figures ~\ref{fig:qwen_pipeline} ~\ref{fig:phi_pipeline} show the results for MMLU Pro. 
We see that the proposed training scheme results in a significant improvement over SFT, curriculum-based SFT, and an alternative training scheme that uses distillation for only easy and medium questions. 
Qwen 3B achieves an accuracy of 0.50 compared to 0.47 and 0.42 for alternative and baseline, respectively, while Phi-4-mini gets to 0.60 compared to 0.58 and 0.46.
At the same time, we see that the pipeline provides comparable performance to the distillation with significantly less data in the number of tokens processed. 
Qwen 3B achieves an accuracy of $0.50$ compared to $0.49$, saving $79\%$ tokens, while Phi-4-mini gets to $0.60$ compared to $0.63$, saving $82\%$ tokens. 
We can also detect an uptrend in the pipeline training, while the full distillation approach exhibits the signs of a plateau. 

To confirm the plateau trend, we ran an extended set of experiments over 20 epochs with the same training split. Table~\ref{tab:results_20epochs} and Figures ~\ref{fig:qwen_pipeline_20epochs} ~\ref{fig:phi_pipeline_20epochs} show the results. 
We find an even stronger confirmation of the previous findings with the proposed scheme outperforming all baselines. 
Qwen 3B achieves an accuracy of $0.52$ compared to $0.50$ and $0.39$ for distillation and baseline, respectively, while Phi-4-mini gets to $0.64$ compared to $0.63$ and $0.51$. 
The same data savings of $79\%$ and $82\%$ apply.

To confirm that the method generalizes to other datasets and models, we perform the same experiments for GSM8K and MedMCQA datasets across two models: LLama 3.2 3B and Phi-4 mini. Table~\ref{tab:results_20epochs} shows the results.
We see that across both models and datasets, our pipeline achieves accuracy either comparable to or surpassing the distillation baseline, with significant data savings.

Training hyperparameters are provided in the Appendix \ref{sec:hyperparameters}.

\begin{table}[ht]
\centering
\begin{tabular}{l l l}
\hline
Method & Qwen 3B & Phi4-mini \\
\hline
SFT & 0.42 / 29k & 0.55 / 27k \\
Curriculum & 0.45 / 29k & 0.54 / 27k \\
Distillation & \underline{0.49} / 1.97m & \textbf{0.63} / 1.51m \\
Alternative & 0.47 / 1.5m & 0.58 / 1.2m  \\
Pipeline (ours) & \textbf{0.50} / \textbf{399k} & \underline{0.60} / \textbf{268k}  \\
\hline
\end{tabular}
\caption{Accuracy / tokens processed for complexity-aware fine-tuning pipelines after 10 epochs (MMLU Pro)}
\label{tab:results_10epochs}
\end{table}

\begin{table*}[ht]
\centering
\begin{tabular}{l cc cc cc}
\toprule
& \multicolumn{2}{c}{MMLU Pro} & \multicolumn{2}{c}{GSM8K} & \multicolumn{2}{c}{MedMCQA} \\
\cmidrule(lr){2-3} \cmidrule(lr){4-5} \cmidrule(lr){6-7}
Method & Qwen 3B & Phi4-mini & LLama 3B & Phi4-mini & LLama 3B & Phi4-mini \\
\midrule
SFT & 0.39 & 0.51 & 0.13 & 0.19 & 0.51 & 0.51 \\
Distillation & \underline{0.50} & \underline{0.63} & \underline{0.79} & \underline{0.89} & \underline{0.56} & \underline{0.57} \\
Pipeline (ours) & \textbf{0.52} & \textbf{0.64} & \textbf{0.82} & \textbf{0.89} & \textbf{0.56} & \textbf{0.58} \\
\bottomrule
\end{tabular}
\caption{Accuracy for complexity-aware fine-tuning pipelines after 20 epochs.}
\label{tab:results_20epochs}
\end{table*}

\subsection{Sensitivity study}
\label{sec:sensitivity_study}

\subsubsection{Alternative complexity estimates}

To analyze the appropriate fine-tuning methods for each complexity band, we first perform the standard SFT for each group.

The same data-splitting procedure described in Section \ref{sec:pipeline_results} is applied to MASJ reasoning score and single-token entropy as complexity metrics.

\begin{figure}[h]
    \centering
    \begin{subfigure}[b]{0.48\linewidth}
        \centering
        \includegraphics[width=\linewidth]{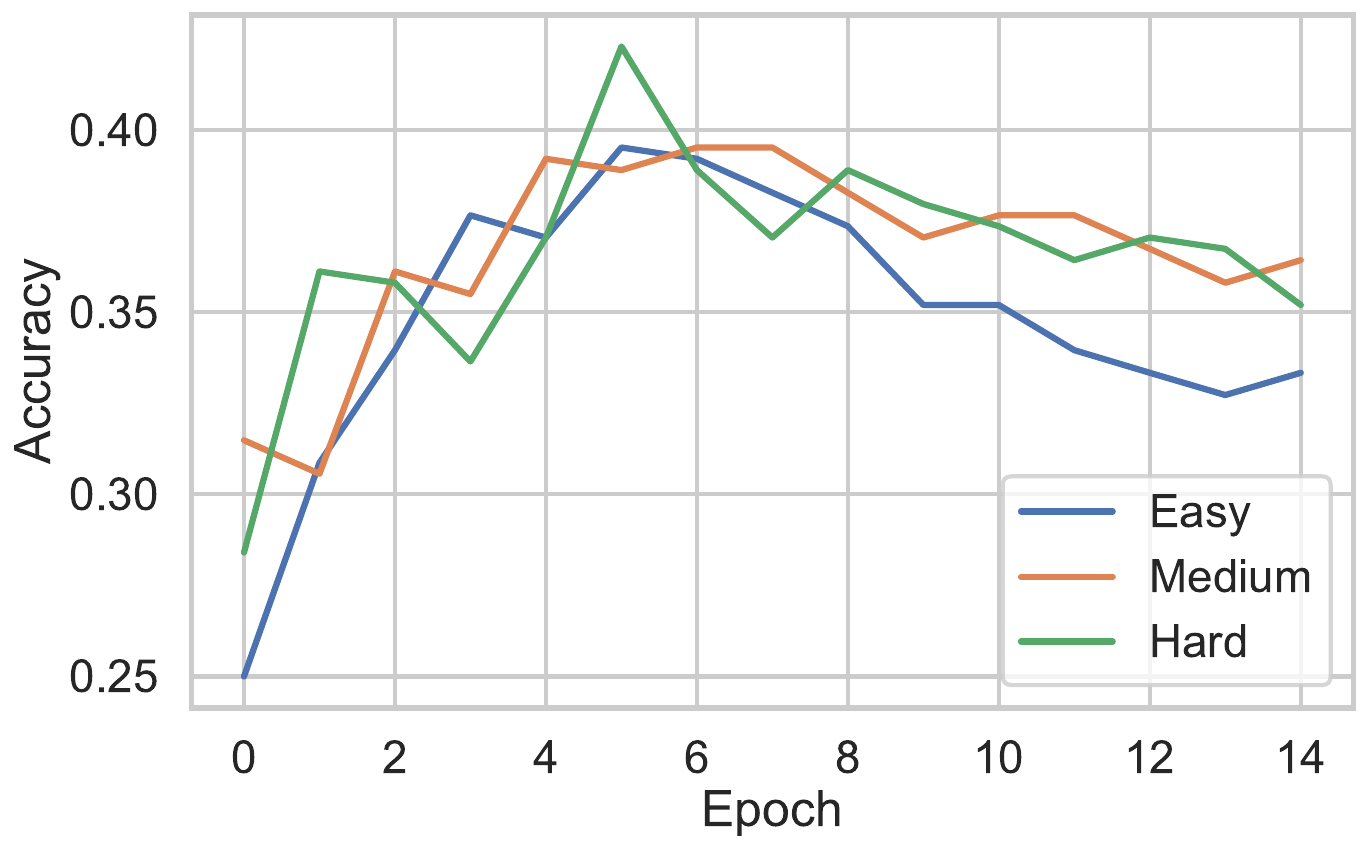}
        \caption{SFT by MASJ reasoning score (Qwen 3B)}
        \label{fig:sft_masj_reasoning_qwen3B}
    \end{subfigure}
    \hfill
    \begin{subfigure}[b]{0.48\linewidth}
        \centering
        \includegraphics[width=\linewidth]{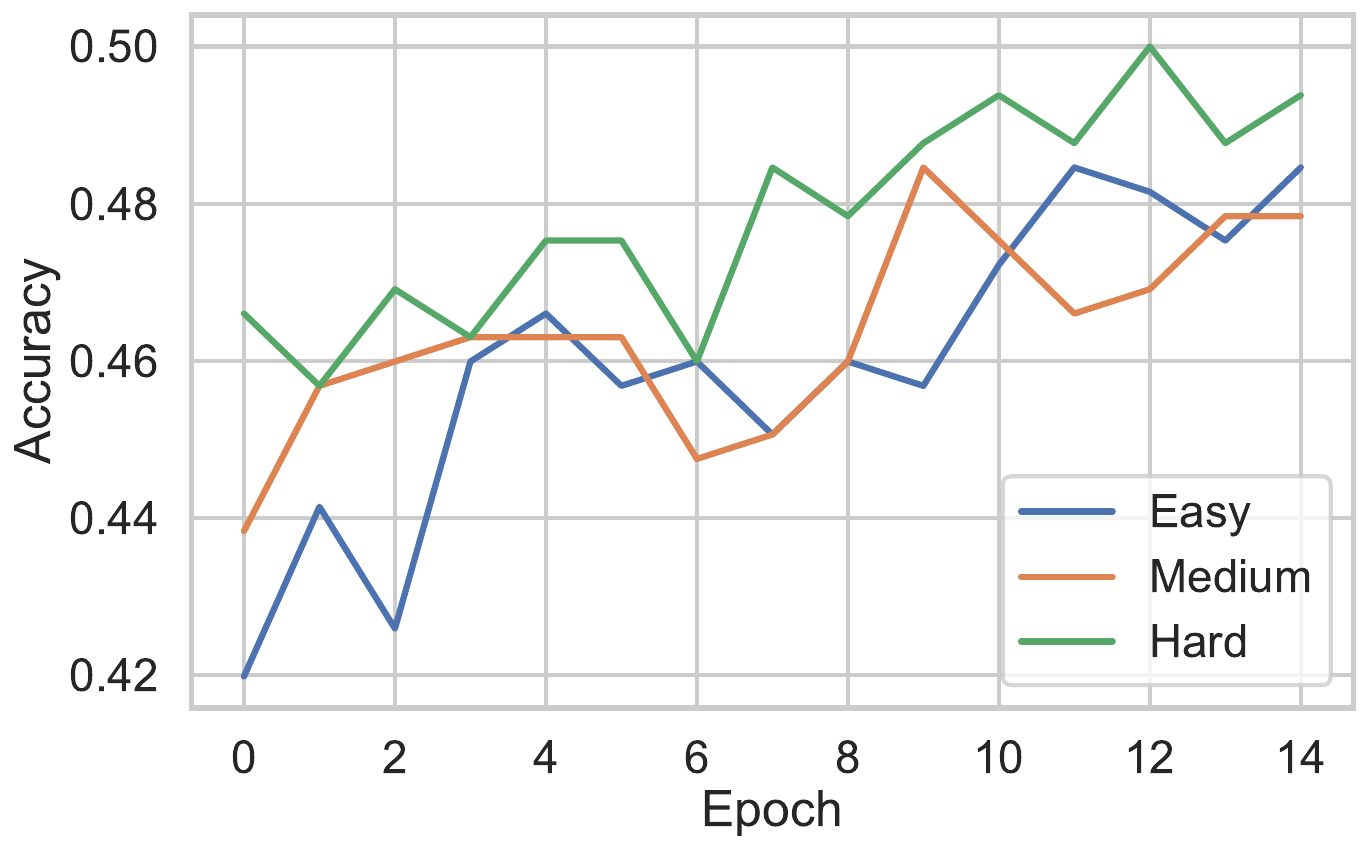}
        \caption{SFT by MASJ reasoning score (Phi-4-mini)}
        \label{fig:sft_masj_reasoning_phi4mini}
    \end{subfigure}
    \hfill
    \begin{subfigure}[b]{0.48\linewidth}
        \centering
        \includegraphics[width=\linewidth]{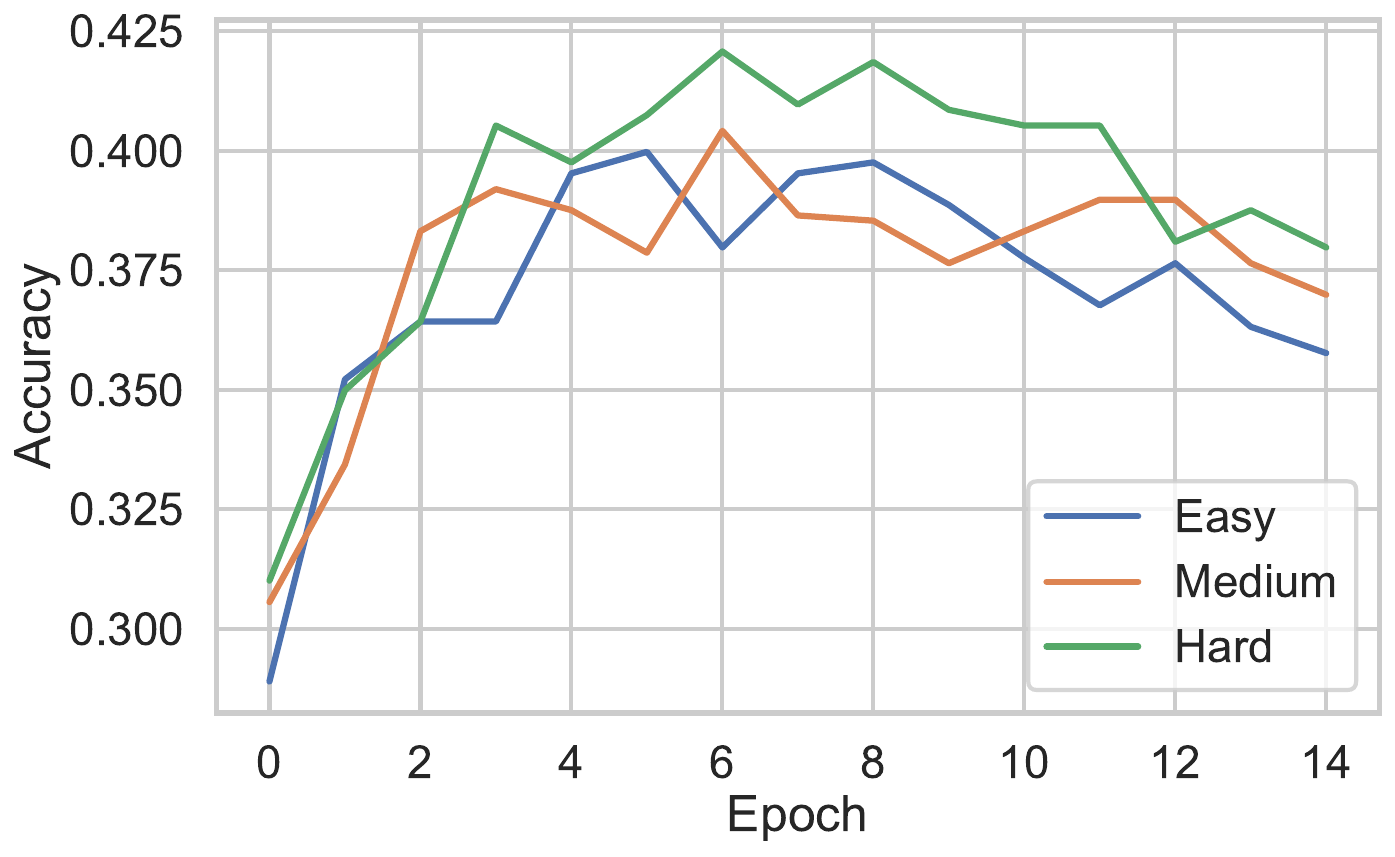}
        \caption{SFT by single token entropy (Qwen 3B)}
        \label{fig:sft_entropy_qwen3B}
    \end{subfigure}
    \hfill
    \begin{subfigure}[b]{0.48\linewidth}
        \centering
        \includegraphics[width=\linewidth]{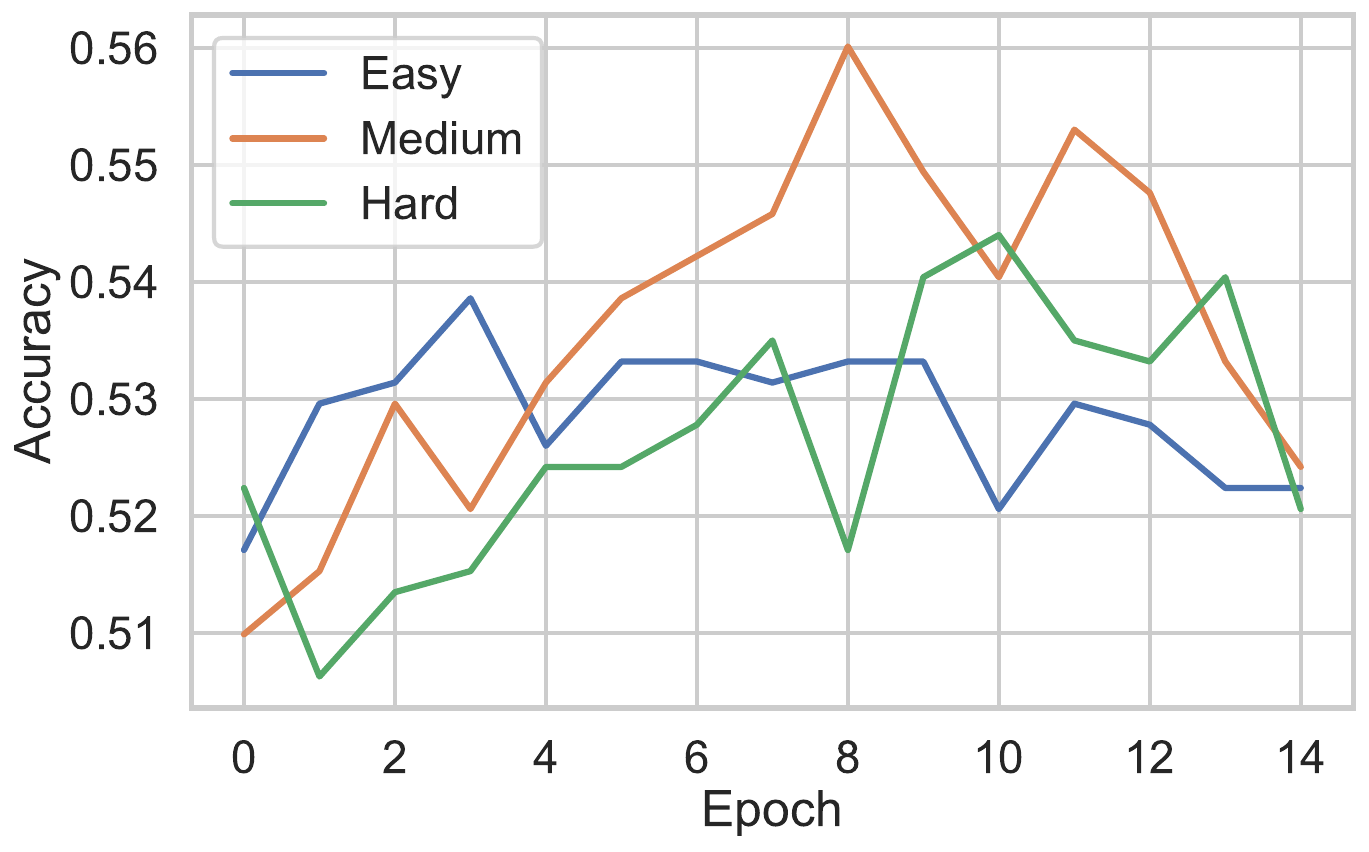}
        \caption{SFT by single token entropy (Phi-4-mini)}
        \label{fig:sft_entropy_phi4mini}
    \end{subfigure}
    \caption{SFT quality dynamics during training with split by complexity estimates provided  by the MASJ reasoning score and the single token entropy across Phi-4-mini and Qwen 3B models.}
    \label{fig:combined_sft_analysis}
\end{figure}

Figures \ref{fig:sft_entropy_phi4mini} \ref{fig:sft_entropy_qwen3B} show the results of SFT for each group split by the single-token entropy. For Phi-4-mini, we see that medium and easy questions outperform hard ones for the first 10 epochs. Shortly after, performance starts to decline for all groups.
For Qwen 3B, we do not see a significant difference between the groups. Moreover, the performance plateaus after five epochs.

Figures \ref{fig:sft_masj_reasoning_phi4mini} and \ref{fig:sft_masj_reasoning_qwen3B} show the results of SFT for each group split by the MASJ reasoning score. We do not see a strong difference in performance between the groups. In combination with questionable ROC AUC scores, provided in Table~\ref{tab:roc_auc_masj}, it makes MASJ reasoning score a less favorable metric for further experiments.

\begin{figure}[h]
    \centering
    \begin{subfigure}[b]{0.48\linewidth}
        \centering
        \includegraphics[width=\linewidth]{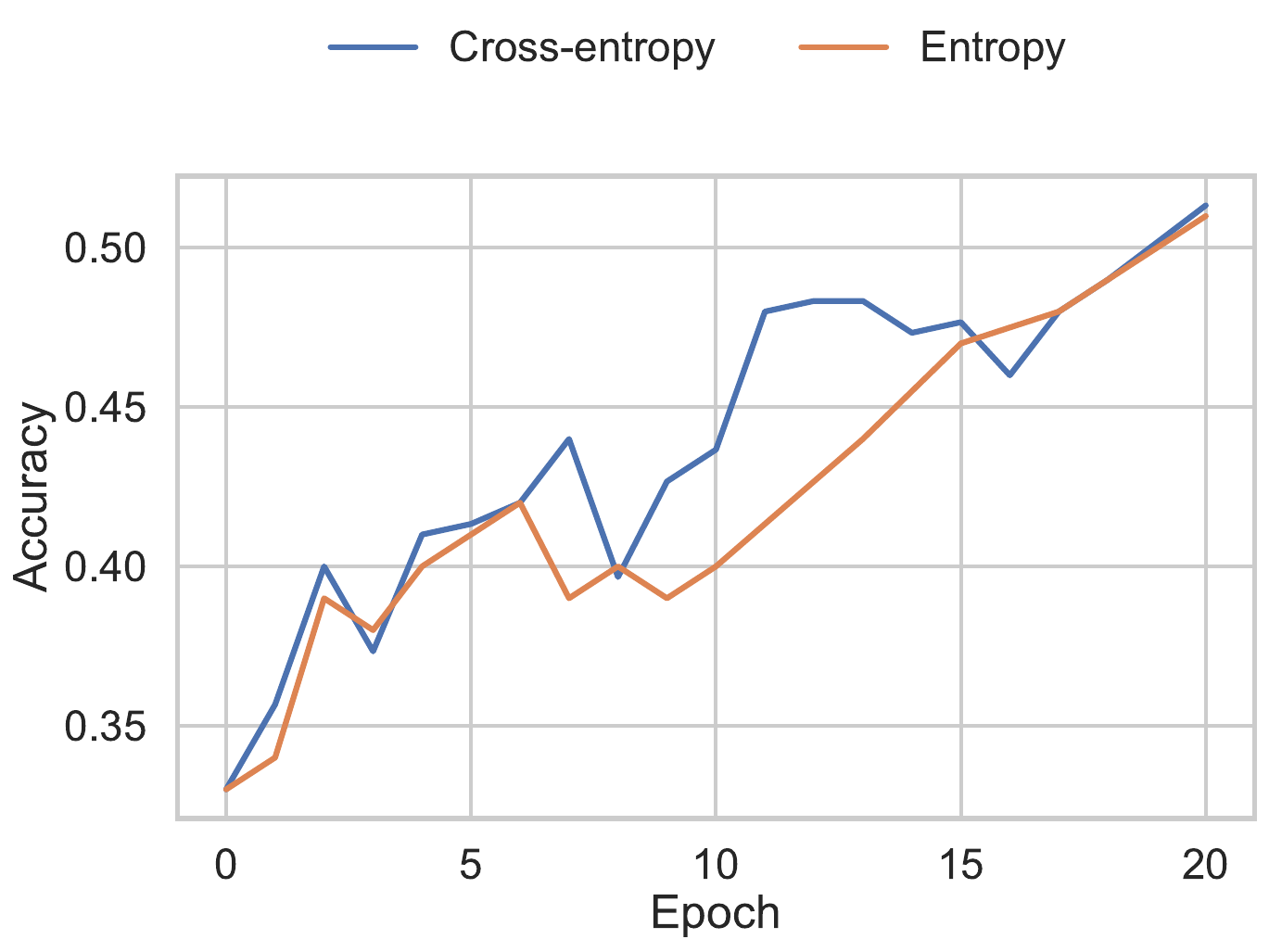}
        \caption{Pipeline performance with cross-entropy (Qwen 3B)}
        \label{fig:entropy_vs_crossentropy_qwen3B}
    \end{subfigure}
    \hfill
    \begin{subfigure}[b]{0.48\linewidth}
        \centering
        \includegraphics[width=\linewidth]{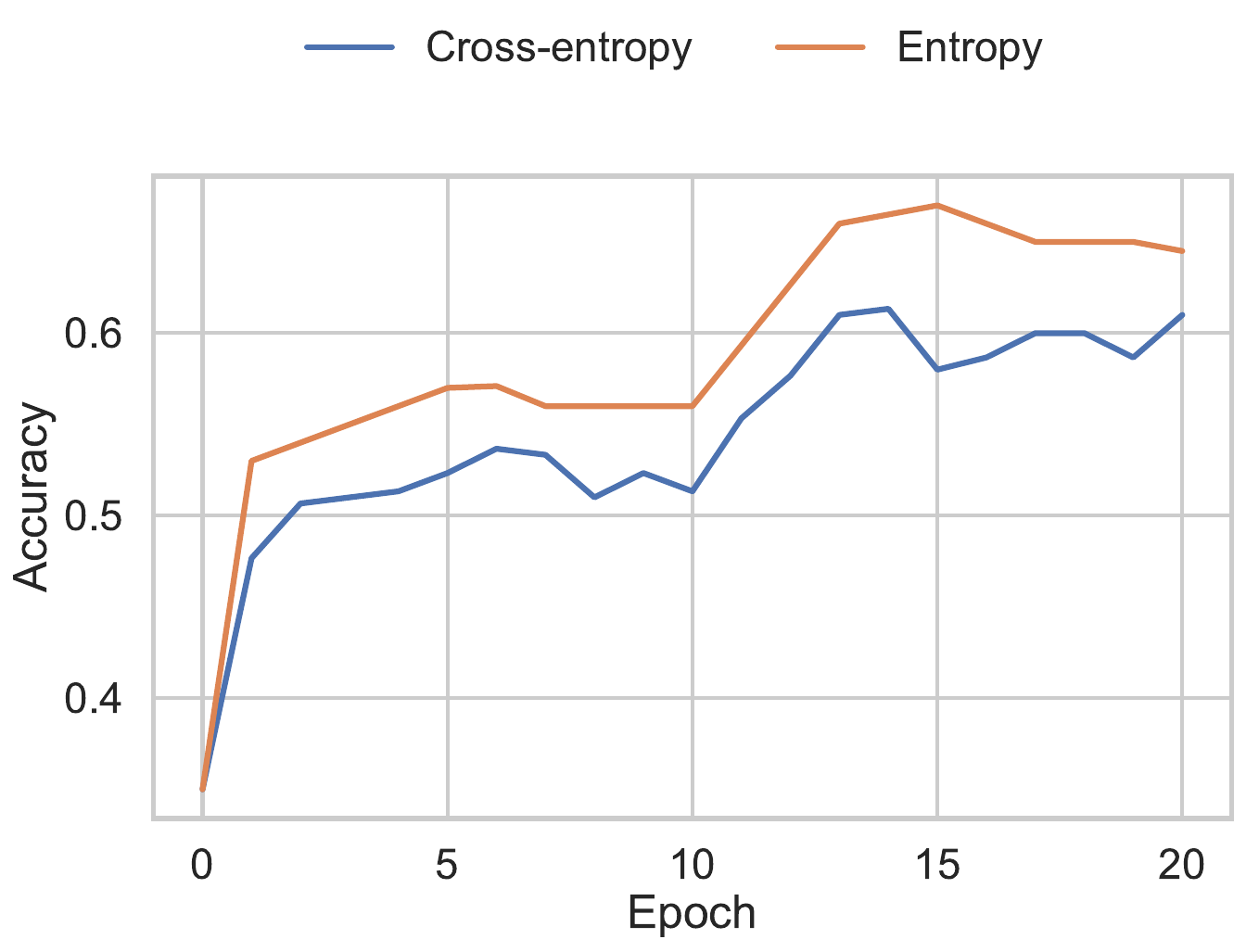}
        \caption{Pipeline performance with cross-entropy (Phi-4-mini)}
        \label{fig:entropy_vs_crossentropy__phi4mini}
    \end{subfigure}
    \caption{Pipeline complexity metric performance comparison for entropy and cross-entropy across Phi-4-mini and Qwen 3B models.}
    \label{fig:entropy_vs_crossentropy_analysis}
\end{figure}

Figures \ref{fig:entropy_vs_crossentropy_qwen3B} and \ref{fig:entropy_vs_crossentropy__phi4mini} show pipeline training results with cross-entropy as a complexity metric compared to entropy. We observe comparable results for Qwen 3B, while Phi4-mini demonstrates superior performance with entropy. The difference in performance could be attributed to entropy capturing the information across the entire distribution instead of a single correct token.

\begin{figure}[h]
    \centering
    \begin{subfigure}[b]{0.48\linewidth}
        \centering
        \includegraphics[width=1\linewidth]{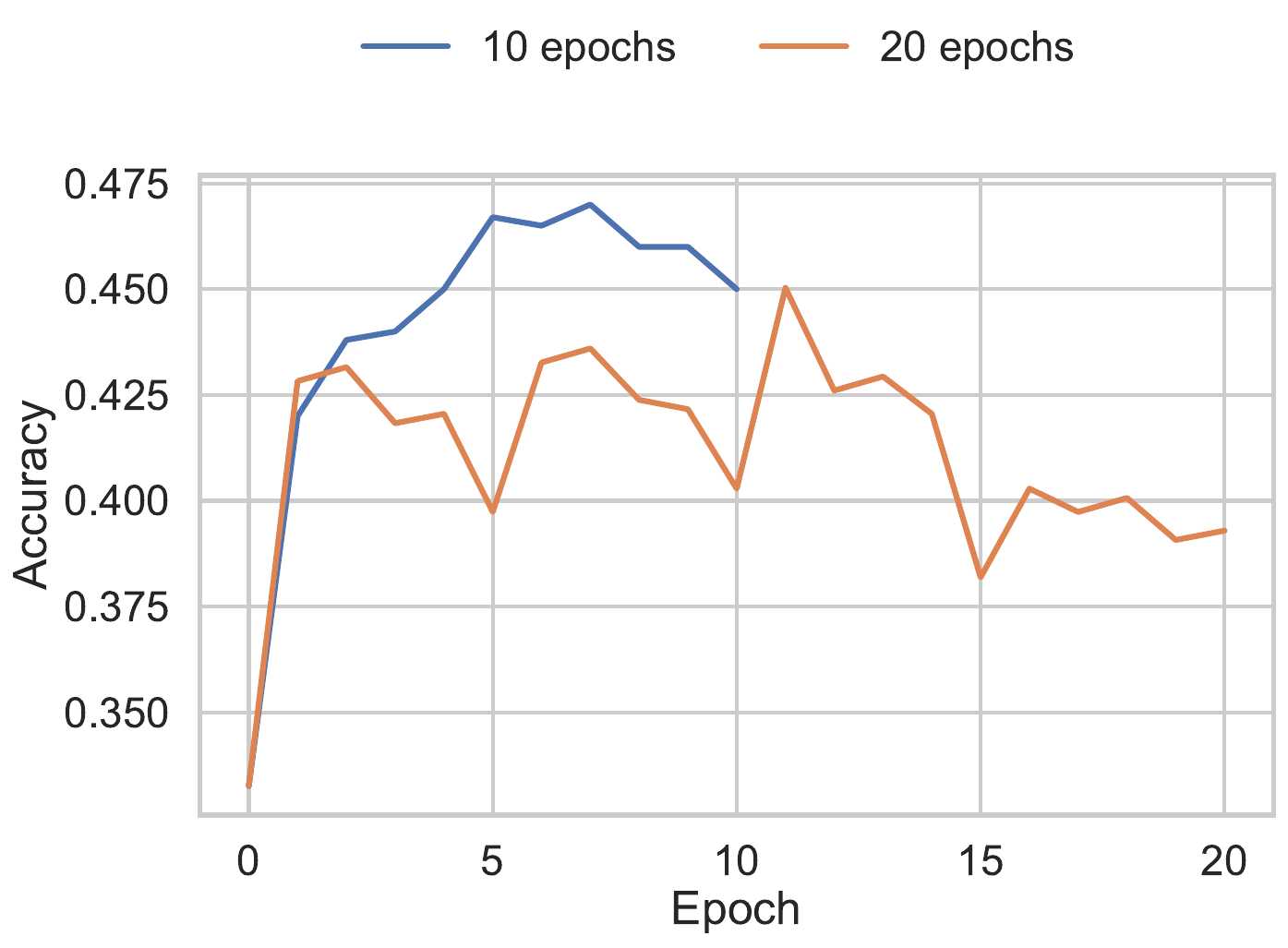}
    \end{subfigure}
    \hfill
    \begin{subfigure}[b]{0.48\linewidth}
        \centering
        \includegraphics[width=1\linewidth]{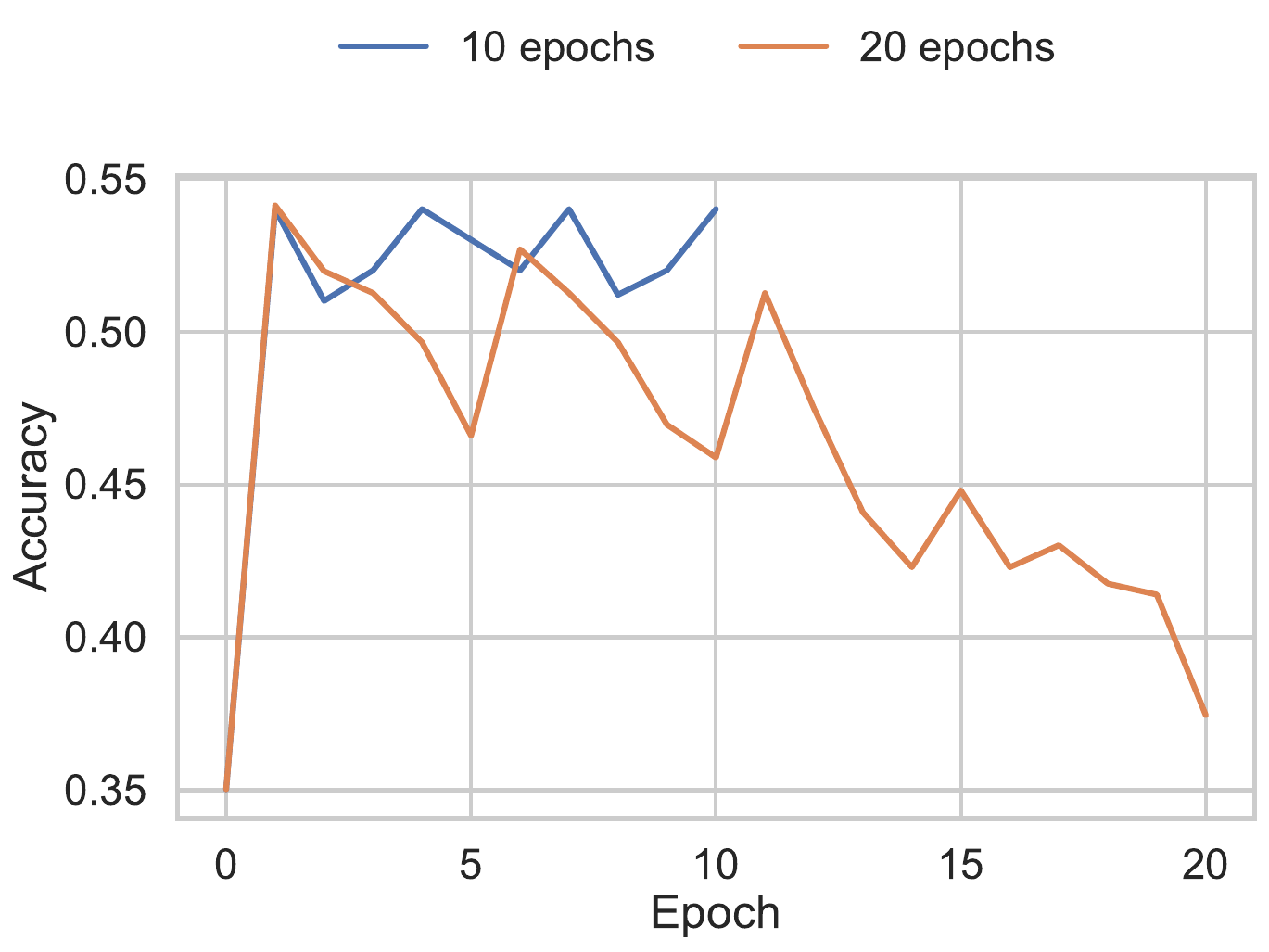}
    \end{subfigure}
    \caption{Curriculum learning accuracy dynamics for different models for Qwen 3B (left) and Phi-4-mini (right)}
    \label{fig:curriculum_plots}
\end{figure}

\begin{table*}[h]
    \centering
    \begin{subtable}[t]{0.48\linewidth}
        \centering
        \begin{tabular}{l c c c c}
            \hline
            Method & Qwen 3B & Phi4-mini \\
            \hline
            Without split & \textbf{0.72} / 0.70 & 0.72 / \textbf{0.74} \\
            \hline
            Education level \\
            \hline
            High school and easier & 0.73 / 0.72 & 0.76 / 0.75 \\
            Undergraduate & 0.73 / 0.71 & 0.72 / 0.77 \\
            Graduate & 0.66 / 0.65 & 0.64 / 0.68 \\
            Postgraduate & 0.63 / 0.52 & 0.64 / 0.63 \\
            \hline
            MASJ reasoning score \\
            \hline
            Low & 0.72 / 0.71 & 0.78 / 0.79 \\
            Medium & 0.72 / 0.70 & 0.70 / 0.72 \\
            High & 0.64 / 0.62 & 0.59 / 0.58 \\
            \hline
        \end{tabular}
        \caption{ROC AUC values for single-token response}
    \end{subtable}
    \hfill
    \begin{subtable}[t]{0.48\linewidth}
        \centering
        \begin{tabular}{l c c c c}
            \hline
            Method & Qwen 3B & Phi4-mini \\
            \hline
            Answer Entropy (AE) & 0.68 / 0.67 & 0.61 / 0.58 \\
            COT Mean & 0.59 / 0.58 & 0.59 / 0.63 \\
            COT Max & 0.63 / 0.61 & 0.6 / 0.65 \\
            Seq Mean & 0.6 / 0.59 & 0.6 / 0.62 \\
            Seq Max-Mean & 0.59 / 0.58 & 0.59 / 0.61 \\
            Seq Mean-Max & 0.62 / 0.6 & 0.59 / 0.62 \\
            Marg Diff Mean & 0.58 / 0.57 & 0.58 / 0.61 \\
            Top-2 Diff & 0.51 / 0.5 & 0.5 / 0.51 \\
            COT Max - AE & 0.54 / 0.53 & 0.51 / 0.57 \\
            COT Max + AE & 0.7 / 0.69 & 0.62 / 0.62 \\
            \hline
        \end{tabular}
        \caption{ROC AUC values for CoT response}
    \end{subtable}
    \caption{ROC AUC values for single-token and CoT responses}
    \label{tab:roc_auc}
    \footnotesize{Second result is for the alternative prompt to allow model answer "I do not know"}\\
\end{table*}

\subsubsection{Curriculum learning fragility}

Multi-stage pipelines such as curriculum learning show extreme sensitivity to hyperparameters such as the number of training epochs, learning rates, etc.

Figure \ref{fig:curriculum_plots} reveals the fragility of the original curriculum learning approach based on SFT. In this experiment, we train a model on the data with increasing complexity for 10 and 20 epochs with different splits: 
\begin{itemize}
    \item 10 epochs: 3 epochs on easy data, 3 epochs on medium data, and 4 epochs on hard data.
    \item 20 epochs: 5 epochs on easy data, 5 epochs on medium data, and 10 epochs on hard data. 
\end{itemize}
Qwen 3B plateaus at a lower accuracy with more training epochs, while Phi4-mini quickly overfits on the data, and its performance degrades.

\subsubsection{Quality of complexity estimation}

We consider three families of uncertainty estimation methods: MASJ, single-token entropy-based, and entropy-based augmented with a chain-of-thought.
MASJ results, as they are inferior to others, are provided in Appendix~\ref{sec:masj_evaluation}.

\paragraph{Single token and chain-of-thought answer entropy}

Table \ref{tab:roc_auc} presents the main ROC AUC values for single token entropy response and for the various aggregates of the chain-of-thought type of response. IDK responses and results with invalid formatting are excluded from the calculations. 

We see that the single-token response yields the best results. In-depth analysis of ROC AUC and accuracy scores with splits across domains and a larger selection of models can be found in tables \ref{tab:roc_auc_single_token_entropy} and \ref{tab:roc_auc_cot_entropy} in the Appendix.

Key observations include:
\begin{itemize}
    \item Single token response provides the best ROC AUC score. At the same time, 'IDK' responses do not consistently affect ROC AUC for all models.
    \item Accuracy tends to be slightly higher when an LLM can answer 'IDK'.
    \item Chain-of-thought responses tend to provide higher accuracy, but lower ROC AUC scores, which makes them less suitable for complexity estimation.
    \item Maximum-based measures (COT Max, Seq Mean Max) consistently outperform mean-based approaches (COT Mean, Seq Max Mean and Seq Mean Mean), suggesting peak uncertainty moments may better indicate question difficulty than average uncertainty.
    \item The poor close-to-random performance of the difference in top entropies suggests that modern LLMs maintain relatively stable reasoning to outliers.
    \item Sequence-based methods did not show good improvements over basic aggregation, indicating that modeling the reasoning structure provides marginal benefits.
\end{itemize}

% \begin{table*}
% \centering
% \begin{tabular}{l c c c c}
% \hline
% Method & Qwen 3B & Qwen 3B* & Phi4-mini & Phi4-mini* \\
% \hline
% Answer Entropy & 0.68/0.36 & 0.67/0.34 & 0.61/0.22 & 0.58/0.17 \\
% COT Mean & 0.59/0.19 & 0.58/0.16 & 0.59/0.19 & 0.63/0.26 \\
% COT Max & 0.63/0.27 & 0.61/0.23 & 0.6/0.2 & \textbf{0.65}/0.3 \\
% Max COT and Answer Entropy Diff & 0.54/0.09 & 0.53/0.06 & 0.51/0.03 & 0.57/0.15 \\
% Sequence Mean Mean & 0.6/0.2 & 0.59/0.17 & 0.6/0.2 & 0.62/0.25 \\
% Sequence Max Mean & 0.59/0.18 & 0.58/0.16 & 0.59/0.18 & 0.61/0.22 \\
% Sequence Mean Max & 0.62/0.25 & 0.6/0.2 & 0.59/0.18 & 0.62/0.24 \\
% Marginal Diff Mean & 0.58/0.17 & 0.57/0.14 & 0.58/0.16 & 0.61/0.23 \\
% Top-2 Entropies Diff & 0.51/0.02 & 0.5/0 & 0.5/0 & 0.51/0.01 \\
% Hybrid Method & \textbf{0.7}/0.41 & \textbf{0.69}/0.37 & \textbf{0.62}/0.24 & 0.62/0.23 \\
% \hline
% Number of Samples & 11049 & 10724 & 9997 & 9973 \\
% \hline

% \end{tabular}
% \caption{ROC AUC / Gini results for COT response}
% \label{tab:roc_auc_cot_aggr_entropy}
% \footnotesize{* Alternative prompt to allow model answer "I do not know"}\\
% \end{table*}

\section{Acknowledgments}

The research was supported by the Russian Science Foundation grant No. 25-11-00355, \url{https://rscf.ru/project/25-11-00355/}. 

\section{Conclusion and discussion}

This paper introduces a complexity-aware fine-tuning pipeline that measures model response uncertainty using the entropy of its predicted answer, then trains on the resulting easy, medium, and hard splits via different tactics.

We confirm that entropy works as a difficulty estimation. Single-token answer entropy reaches ROC AUC values up to $0.8$, clearly beating MASJ-based estimates of $0.57$. 
This confirms that a model's own confidence is a reliable, automatic proxy for question difficulty.

Using the entropy-based data split, we find that different complexity scores require different training approaches. Standard supervised fine-tuning (SFT) is enough for the easy and medium bands, but lags on the hard band. For hard questions, adding a distilled chain of thought from a large LLM unlocks further gains. Our pipeline achieves accuracies of 0.52/0.64 vs. 0.39/0.51 for Qwen 3B/Phi-4-mini, while using $81\%$ less data.

The pipeline is fully automated and can be included in other fine-tuning workflows. Our findings suggest that curriculum ideas still matter for today's LLMs: letting the model focus on what it can already solve directly, while giving extra guidance only where it struggles, yields a better allocation of limited model capacity.

\section*{Limitations}

\begin{itemize}
    \item The proposed pipeline is tested only on small models. Results may differ across other question-answering datasets, open-ended tasks, other domains, or larger LLMs.
    \item In low-resource settings teacher may be unavailable or imperfect, which reduces the benefit of learning from a distilled chain-of-thought. Additionally, we did not explore how well the approach generalizes to other reasoning-promoting techniques.
    \item Low entropy can still correspond to hallucinations, which leads to imperfect identification of the question complexity.
    \item We split the data into 3 equal parts and did not explore other possible boundaries.
    \item We did not conduct an extensive ablation study, which might reveal that our approach does not yield the best possible training combination or sequence within the current framework. It remains an area for further research.
\end{itemize}

% Bibliography entries for the entire Anthology, followed by custom entries
%\bibliography{anthology,custom}
% Custom bibliography entries only
\bibliography{custom}

\clearpage

\appendix
\section{Prompts}
\label{sec:prompts}

\subsection{Prompts used for complexity estimation}
\label{sec:uq_prompts}

\begin{table}[H]
\small
\centering
\begin{tabularx}{\linewidth}{X}
\hline 
\emph{System prompt for a single token response} \\
\\
The following are multiple choice questions about {subject}. Write down ONLY the NUMBER of the correct answer and nothing else.
\\
\hline 
\emph{System prompt for a single token response with a fallback for unknown answers} \\
\\
The following are multiple choice questions about {subject}. If you are certain about the answer return the correct option number, otherwise return 0. Write down ONLY the NUMBER and nothing else. \\
\hline 
\emph{System prompt for a single token response with a fallback for unknown answers (alternative)} \\
\\
The following are multiple choice questions about {subject}. If you know the answer return the correct option number, otherwise return 0. Write down ONLY the NUMBER and nothing else.
\\
\hline 
\emph{System prompt for a chain-of-thought response} \\
\\
The following are multiple choice questions about {subject}. Explain your thinking process step-by-step. At the end, write down the number of the correct answer by strictly following this format: $\lbrack\lbrack\text{number of correct answer}\rbrack\rbrack$.
\\
\hline 
\emph{System prompt for a chain-of-thought response with a fallback for unknown answers} \\
\\
The following are multiple choice questions about {subject}. Explain your thinking process step-by-step. At the end, if you are certain about the answer write down the number of the correct answer by strictly following this format: $\lbrack\lbrack\text{number of correct answer}\rbrack\rbrack$, otherwise return $\lbrack\lbrack\text{0}\rbrack\rbrack$.
\\
\hline 
\emph{System prompt for a chain-of-thought response with a fallback for unknown answers (alternative)} \\
\\
The following are multiple choice questions about {subject}. Explain your thinking process step-by-step. At the end, if you know the answer write down the number of the correct answer by strictly following this format: $\lbrack\lbrack\text{number of correct answers}\rbrack\rbrack$, otherwise return $\lbrack\lbrack\text{0}\rbrack\rbrack$.
\\
\hline 
\emph{User prompt} \\
Question: ...\\
Options:\\
1. ...\\
2. ...\\
...\\
n. ...\\
Choose one of the answers. Write down ONLY the NUMBER of the correct answer and nothing else.
\\
\hline
\end{tabularx}
\caption{Prompts for complexity estimation}
\label{tab:prs_uqs_uq}
\end{table}

\subsection{General prompts}

\begin{table}[H]
\small
\centering
\begin{tabularx}{\linewidth}{X}
\hline
You are an expert in the topic of the question. Please act as an impartial judge and evaluate the complexity of the multiple-choice question with options below.
Begin your evaluation by providing a short explanation. Be as objective as possible. After providing your explanation, you must not answer the question.
You must rate the question complexity by strictly following the criteria:
[[Number of reasoning steps]] - how many reasoning steps do you need to answer this question? Valid answers: low, medium, high.
Your answer must strictly follow this format: "[[Number of reasoning steps: answer]]".
Example 1: "Your explanation... [[Number of reasoning steps: low]]".
Example 2: "Your explanation... [[Number of reasoning steps: high]]".
Example 3: "Your explanation... [[Number of reasoning steps: medium]]".
\\
\hline 
\end{tabularx}
\caption{Prompt for MASJ reasoning}
\label{tab:prompt_masj_reasoning}
\end{table}

\begin{table}[H]
\small
\centering
\begin{tabularx}{\linewidth}{X}
\hline
You are an expert in the topic of the question. Please act as an impartial judge and evaluate the complexity of the multiple-choice question with options below. Begin your evaluation by providing a short explanation. Be as objective as possible. After providing your explanation, you must not answer the question. You must rate the question complexity by strictly following the scale: "high school and easier", "undergraduate", "graduate", "postgraduate". You must return the complexity by strictly following this format: "[[complexity]]", for example: "Your explanation... Complexity: [[undergraduate]]", which corresponds to the undergraduate level.
\\
\hline 
\end{tabularx}
\caption{Prompt for MASJ education levels}
\label{tab:prompt_masj_edu_levels}
\end{table}

\section{Aggregation Methods}
\label{sec:aggr_methods}

Our analysis considers diverse methods to identify the complexity of a questions based on outputs logits with a specific focus on commonly used entropy.
They are listed in the list with more details provided below:
\begin{enumerate}
    \item CoT word-aggregation methods
    \begin{itemize}
        \item Single Token Answer Entropy
        \item CoT Mean
        \item CoT Max
        \item Difference between CoT Max and Single Token Answer Entropy
    \end{itemize}
    \item CoT sequence-aggregation methods
    \begin{itemize}
        \item Sequence Mean of Words Mean 
        \item Sequence Max of Words Mean
        \item Sequence Mean of Words Max
    \end{itemize}
    \item Probability-based methods
    \begin{itemize}
        \item Mean of Marginal Difference - mean of difference between top-2 probabilities for each token of response
        \item Top-2 Entropy Difference - difference of top-2 highest entropies for response
    \end{itemize}
    \item Hybrid method
    \begin{itemize}
        \item Mix of CoT word-aggregation methods - linear combination of the best perform methods
    \end{itemize}
\end{enumerate}

\subsection{Word-aggregation Methods}
This CoT aggregations have the same entropy values as in Section \ref{sec:method_single_token_entropy} for each CoT token:
$$
h_j = -\sum_{i=1}^{n} p_i \log{p_i},
$$
where $p_i$ is the probability of a specific token, $n$ is the vocabulary size, and $h_j$ is the entropy of the corresponded token.
Aggregating per-token entropies for an answer of length $N$ get:
\begin{align*}
\mathrm{CoT}_{\mathrm{mean}} &= \frac{1}{N} \sum_{j=1}^{N} h_j, \\
\mathrm{CoT}_{\mathrm{max}} &= \max_j h_j.
\end{align*}
So, Chain-of-Thought maximum and answer entropy difference is:
$$
|\mathrm{CoT}_{\mathrm{max}}  - h_{\mathrm{answer}}|,
$$
where $h_{\mathrm{answer}}$ is the entropy of the answer token.

\subsection{Sequence-aggregation Methods}
For $M$ logical claims, which were split by tokens that corresponded to the end of the sequence, we have tokens sets $C_1, C_2, \ldots, C_M$.
\begin{align*}
\mathrm{Seq}_{\mathrm{mean}} &= \frac{1}{M} \sum_{j=1}^M \left[\frac{1}{|C_j|} \sum_{i \in C_j} h_i \right], \\
\mathrm{Seq}_{\mathrm{mean, max}} &= \frac{1}{M} \sum_{j=1}^M \left[ \max_{i \in C_j} h_i \right], \\    
\mathrm{Seq}_{\mathrm{max, mean}} &= \max_j \left[ \frac{1}{|C_j|} \sum_{i \in C_j} h_i \right].
\end{align*}

\subsection{Probability-based Methods}
Assume that for each token in response, we have the token probability distribution $p_i$. So, the marginal token difference is $\sigma_i$ and mean marginal difference is mean of all $\sigma_i$ in LLM response.
\begin{align*}
\sigma_i &= p_i^(1) - p_i^(2), \\   
\overline{\sigma} &= \frac{1}{N} \sum_{i=1}^N \sigma_i,
\end{align*}
here $p_i^(k)$ is the $k$-th top value in the vector $\mathbf{p}_i$ of the probability distribution of the tokens from the dictionary during the generation of $i$-th token.

We can also consider differences between top two entropies in response $\delta$:
\[
\delta = \max_{j} h_j - \max_{i | i \ne j} h_i.
\]

\subsection{Hybrid Method}
We can also use a linear combination of $2$ best-perform previous methods: $h_{\mathrm{answer}}$ and $\mathrm{CoT}_{\max}$. Also, we tried adding the third element $\mathrm{CoT}_{\mathrm{mean}}$, but it has decreased the ROC-AUC, so we made a decision to remove it.
\[
h_{\mathrm{mix}} = (1 - \alpha) h_{\mathrm{answer}} + \alpha \mathrm{CoT}_{\mathrm{max}},
\]
where $0 \leq \alpha \leq 1$ is the hyperparameter.
Empirically we identified that the best value for $\alpha$ is $0.05$ for Qwen-3B.

\subsection{Detailed Results}

\begin{table*}
\centering
\resizebox{\textwidth}{!}{%
\begin{tabular}{l c c c c c c c c}
\hline
Category & Qwen 3B & Qwen 3B* & Phi4-mini & Phi4-mini* & Phi4 & Phi4* & Mistral 24B & Mistral 24B* \\
\hline
All & 0.72/0.33 & 0.70/0.33 & 0.72/0.40 & 0.74/0.46 & \textbf{0.80}/0.51 & \textbf{0.80}/0.58 & 0.75/0.49 & 0.74/0.60 \\
\hline
Law & 0.63/0.24 & 0.60/0.21 & 0.64/0.29 & 0.62/0.30 & 0.69/0.47 & 0.69/0.48 & 0.69/0.41 & \textbf{0.75}/0.56 \\
Business & 0.67/0.28 & 0.71/0.26 & 0.67/0.31 & 0.64/0.38 & 0.73/0.36 & \textbf{0.75}/0.44 & 0.69/0.40 & 0.68/0.43 \\
Psychology & 0.77/0.51 & 0.75/0.51 & \textbf{0.84}/0.57 & 0.82/0.59 & \textbf{0.84}/0.74 & \textbf{0.84}/0.74 & 0.79/0.66 & 0.75/0.68 \\
Chemistry & 0.69/0.23 & 0.62/0.24 & 0.62/0.34 & 0.64/0.41 & 0.70/0.34 & \textbf{0.77}/0.45 & 0.68/0.38 & 0.75/0.59 \\
Biology & 0.79/0.59 & 0.79/0.56 & 0.85/0.67 & 0.85/0.73 & \textbf{0.90}/0.80 & \textbf{0.90}/0.83 & 0.81/0.74 & 0.73/0.80 \\
History & 0.66/0.36 & 0.63/0.35 & 0.68/0.39 & 0.65/0.43 & \textbf{0.76}/0.62 & 0.73/0.63 & 0.69/0.54 & 0.64/0.56 \\
Other & 0.70/0.33 & 0.67/0.34 & 0.72/0.39 & 0.74/0.43 & 0.81/0.57 & \textbf{0.82}/0.58 & 0.79/0.52 & 0.75/0.59 \\
Physics & 0.65/0.27 & 0.64/0.28 & 0.65/0.32 & 0.66/0.40 & 0.75/0.39 & \textbf{0.78}/0.46 & 0.74/0.38 & 0.71/0.63 \\
Computer science & 0.76/0.29 & 0.70/0.32 & 0.73/0.41 & 0.76/0.46 & 0.77/0.55 & \textbf{0.80}/0.57 & 0.77/0.51 & 0.74/0.64 \\
Health & 0.69/0.39 & 0.66/0.39 & 0.71/0.43 & 0.71/0.47 & \textbf{0.78}/0.64 & 0.77/0.65 & 0.75/0.61 & 0.71/0.63 \\
Economics & 0.77/0.44 & 0.74/0.43 & 0.79/0.55 & 0.80/0.59 & \textbf{0.85}/0.68 & 0.83/0.72 & 0.77/0.62 & 0.75/0.66 \\
Math & 0.69/0.24 & 0.67/0.24 & 0.65/0.27 & 0.69/0.31 & 0.73/0.37 & \textbf{0.74}/0.43 & 0.69/0.33 & 0.72/0.44 \\
Philosophy & 0.66/0.33 & 0.70/0.31 & 0.71/0.39 & 0.70/0.43 & \textbf{0.77}/0.61 & 0.76/0.63 & 0.71/0.53 & 0.70/0.56 \\
Engineering & 0.67/0.34 & 0.66/0.32 & 0.62/0.39 & 0.64/0.45 & 0.74/0.43 & 0.67/0.53 & 0.70/0.46 & \textbf{0.77}/0.60 \\
\hline
Education level \\
\hline
High school and easier & 0.73/0.35 & 0.72/0.34 & 0.76/0.38 & 0.75/0.51 & 0.81/0.50 & \textbf{0.82}/0.54 & 0.75/0.48 & 0.70/0.52 \\
Undergraduate & 0.73/0.34 & 0.71/0.34 & 0.72/0.42 & 0.77/0.44 & 0.81/0.52 & \textbf{0.82}/0.62 & 0.77/0.50 & 0.74/0.64 \\
Graduate & 0.66/0.28 & 0.65/0.26 & 0.64/0.35 & 0.68/0.37 & 0.74/0.50 & 0.73/0.54 & 0.71/0.46 & \textbf{0.76}/0.58 \\
Postgraduate & 0.63/0.18 & 0.52/0.20 & 0.64/0.20 & 0.63/0.22 & \textbf{0.67}/0.40 & 0.65/0.41 & 0.62/0.35 & 0.63/0.39 \\
\hline
MASJ reasoning score \\
\hline
Low & 0.72/0.42 & 0.71/0.42 & 0.78/0.48 & 0.79/0.51 & 0.82/0.64 & \textbf{0.83}/0.65 & 0.79/0.59 & 0.73/0.59 \\
Medium & 0.72/0.32 & 0.70/0.31 & 0.70/0.39 & 0.72/0.44 & \textbf{0.79}/0.50 & \textbf{0.79}/0.59 & 0.74/0.47 & 0.76/0.63 \\
High & 0.64/0.27 & 0.62/0.27 & 0.59/0.33 & 0.58/0.36 & \textbf{0.69}/0.41 & 0.64/0.29 & 0.64/0.39 & 0.62/0.45 \\
\hline
\end{tabular}}
\caption{ROC AUC/accuracy for single token response entropy}
\label{tab:roc_auc_single_token_entropy}
\footnotesize{* Alternative prompt to allow model answer "I do not know"}\\
\end{table*}

\begin{table*}
\centering
\begin{tabular}{l c c c c}
\hline
Category & Qwen 3B & Qwen 3B* & Phi4-mini & Phi4-mini* \\
\hline
All & \textbf{0.68}/0.41 & 0.67/0.41 & 0.61/0.43 & 0.58/0.55 \\
\hline
Law & \textbf{0.60}/0.24 & 0.57/0.23 & 0.55/0.26 & 0.52/0.28 \\
Business & \textbf{0.68}/0.45 & 0.67/0.47 & 0.66/0.55 & 0.56/0.65 \\
Psychology & \textbf{0.73}/0.51 & 0.70/0.51 & 0.68/0.48 & 0.65/0.63 \\
Chemistry & 0.65/0.41 & \textbf{0.68}/0.39 & 0.65/0.43 & 0.63/0.60 \\
Biology & \textbf{0.77}/0.56 & 0.68/0.60 & 0.65/0.48 & 0.67/0.71 \\
History & \textbf{0.62}/0.36 & 0.61/0.36 & 0.59/0.37 & 0.51/0.39 \\
Other & \textbf{0.63}/0.38 & \textbf{0.63}/0.36 & 0.60/0.42 & 0.58/0.52 \\
Physics & \textbf{0.68}/0.42 & 0.67/0.41 & 0.62/0.39 & 0.61/0.57 \\
Computer science & 0.68/0.37 & \textbf{0.73}/0.33 & 0.59/0.41 & 0.58/0.58 \\
Health & \textbf{0.63}/0.37 & 0.61/0.40 & 0.62/0.33 & 0.56/0.50 \\
Economics & \textbf{0.70}/0.48 & 0.68/0.50 & 0.61/0.47 & 0.65/0.63 \\
Math & \textbf{0.73}/0.51 & \textbf{0.73}/0.48 & 0.63/0.58 & 0.60/0.67 \\
Philosophy & 0.63/0.33 & 0.62/0.35 & \textbf{0.66}/0.37 & 0.59/0.48 \\
Engineering & 0.63/0.31 & \textbf{0.64}/0.33 & 0.60/0.37 & 0.55/0.45 \\
\hline
Education level \\
\hline
High school and easier & \textbf{0.72}/0.56 & 0.70/0.53 & 0.66/0.57 & 0.60/0.73 \\
Undergraduate & \textbf{0.67}/0.41 & \textbf{0.67}/0.41 & 0.62/0.42 & 0.60/0.55 \\
Graduate & \textbf{0.61}/0.27 & 0.60/0.28 & 0.58/0.30 & 0.57/0.38 \\
Postgraduate & 0.63/0.22 & \textbf{0.66}/0.15 & 0.41/0.22 & 0.41/0.20 \\
\hline
MASJ reasoning score \\
\hline
Low & \textbf{0.69}/0.49 & 0.67/0.49 & 0.64/0.46 & 0.61/0.62 \\
Medium & \textbf{0.67}/0.41 & 0.66/0.41 & 0.60/0.43 & 0.57/0.55 \\
High & \textbf{0.65}/0.26 & 0.60/0.26 & 0.53/0.32 & 0.54/0.36 \\
\hline
\end{tabular}
\caption{ROC AUC/accuracy for single token response entropy after chain-of-thought}
\label{tab:roc_auc_cot_entropy}
\footnotesize{* Alternative prompt to allow model answer "I do not know"}\\
\end{table*}

\section{Additional experiments}
\label{sec:additional_experiments}

\subsection{MASJ education level and reasoning score}
\label{sec:masj_evaluation}

\begin{table}[t]
\centering
\begin{tabular}{l c c}
\hline
Model & Education level & Reasoning \\
\hline
Qwen 3B & \underline{0.53} & 0.55 \\
Qwen 3B* & \underline{0.53} & 0.55 \\
Phi4-mini & 0.52 & 0.55 \\
Phi4-mini* & 0.52 & 0.54 \\
Phi4 & 0.50 & \underline{0.57} \\
Phi4* & 0.50 & 0.55 \\
Mistral 24B & 0.50 & 0.56 \\
Mistral 24B* & 0.52 & 0.53 \\
\hline
\end{tabular}
\caption{ROC AUC for MASJ}
\label{tab:roc_auc_masj}
\footnotesize{* Alternative prompt to allow model answer "I do not know"}\\
\end{table}

Table~\ref{tab:roc_auc_masj} shows ROC AUC values for MASJ evaluations of education levels and reasoning scores.

We can see that MASJ reasoning score has a slightly higher ROC AUC of 0.55 on average compared to education levels with ROC AUC of 0.52. There is no significant difference between prompts that allow IDK answers and the ones that do not. 

The results indicate that MASJ scores divide the data into complexity groups with moderate quality. On the other hand, results depend on encoding of nominal scores provided by MASJ, and a more comprehensive study could improve this method.
% It suggests that MASJ scores might not be the best metrics to split the data into complexity groups. However, ROC AUC calculations are highly sensitive to the encoding of the nominal MASJ scores. It suggests further research of the MASJ reasoning score that shows an edge even with the current uniformly distributed encodings.

\paragraph{Technical details.}
To calculate ROC AUC we encode MASJ results on a scale from 0 to 1 and prompt the model to answer questions directly, using prompts. For education levels, we take "High school and easier" - 0.2, "Undergraduate" - 0.4, "Graduate" - 0.6, "Postgraduate" - 0.8. For reasoning scores, "Low" - 0.25, "Medium" - 0.5, "High" - 0.75.
IDK responses and results with invalid formatting are excluded from the calculations.

\subsection{Feature importances of reasoning model}
\label{sec:reasoner_feat_imp}

We evaluated logistic regression weights, that reflect feature importance. Our classifier achieves $0.721$, $0.717$ and $0.731$ accuracies by using thinking total entropy, length of the reasoning chain or both features combined correspondingly, thus producing a reasonable model for the evaluation of a probability of the error for question.

Table \ref{tab:param_importances} shows that total entropy and number of tokens of the reasoning chain are the most important parameters influencing the correctness of the model's prediction. 

\paragraph{Technical details.}

To avoid excessively long reasoning chains, we set a maximum generation length of 5000 tokens. We also use normalized parameters to remove the mean and scale to unit variance. We take model coefficients of the corresponding parameters as their importance. 

\begin{table}[t]
\centering
\begin{tabular}{l c}
\hline
Statistics & Importance \\
\hline
Thinking total entropy & \underline{1.45}\\
Thinking number of tokens & \underline{1.08}\\
Answer total entropy & 0.25\\
Answer length & 0.20\\
\hline
\end{tabular}
\caption{Absolute values of parameter weights}
\label{tab:param_importances}
\end{table}

\section{Hyperparameters}
\label{sec:hyperparameters}

\begin{table*}[ht]
\centering
\begin{tabular}{l c c c c}
\hline
Parameter & SFT & Pipeline (easy/mid) & Pipeline (hard) & Distillation \\
\hline
Effective batch size & 64 & 64 & 64 & 64\\
Max new tokens & 30 & 30 & 8192 & 8192\\
Optimizer & AdamW & AdamW & AdamW & AdamW \\
Learning Rate & 1e-4 & 1e-4 & 1e-4 & 1e-4\\
\hline
\end{tabular}
\caption{Training hyperparameters (MMLU Pro)}
\label{tab:hypers_mmlu}
\end{table*}

\begin{table*}[ht]
\centering
\begin{tabular}{l c c c c}
\hline
Parameter & SFT & Pipeline (easy/mid) & Pipeline (hard) & Distillation \\
\hline
Effective batch size & 256 & 256 & 256 & 256\\
Max new tokens & 100 & 100 & 8192 & 8192\\
Optimizer & AdamW & AdamW & AdamW & AdamW \\
Learning Rate & 1e-4 & 1e-4 & 1e-4 & 1e-4\\
LoRA rank & 64 & 64 & 64 & 64\\
LoRA alpha & 128 & 128 & 128 & 128\\
LoRA dropout & 0.05 & 0.05 & 0.05 & 0.05\\
\hline
\end{tabular}
\caption{Training hyperparameters (GSM8K/MedMCQA)}
\label{tab:hypers_rest}
\end{table*}

\end{document}